\pdfoutput=1
\documentclass[letter,oneside]{article}
\usepackage[margin=1.0in]{geometry}

\usepackage[numbers]{natbib}

\usepackage{times}  
\usepackage{helvet}  
\usepackage{courier}  
\usepackage[hyphens]{url}  
\usepackage{graphicx} 
\usepackage{amsfonts} 
\usepackage{bbm} 
\usepackage{soul}
\usepackage{natbib}  
\usepackage{caption} 
\usepackage{subcaption}
\usepackage{amsmath} 
\usepackage{booktabs}
\usepackage{multirow}
\usepackage{siunitx}
\usepackage{xcolor}

\usepackage[utf8]{inputenc} 
\usepackage[T1]{fontenc}    
\usepackage{hyperref}       
\usepackage{url}            
\usepackage{booktabs}       
\usepackage{amsfonts}       
\usepackage{nicefrac}       
\usepackage{microtype}      
\usepackage{graphicx}
\usepackage{algorithm2e,algorithmic,amsmath,amssymb,amsthm}
\usepackage{multicol}
\usepackage{multirow}
\usepackage{listings}
\usepackage{xcolor}

\definecolor{codegreen}{rgb}{0,0.6,0}
\definecolor{codegray}{rgb}{0.5,0.5,0.5}
\definecolor{codepurple}{rgb}{0.58,0,0.82}
\definecolor{backcolour}{rgb}{0.95,0.95,0.92}

\lstdefinestyle{mystyle}{
    backgroundcolor=\color{backcolour},   
    commentstyle=\color{codegreen},
    keywordstyle=\color{magenta},
    numberstyle=\tiny\color{codegray},
    stringstyle=\color{codepurple},
    basicstyle=\ttfamily\footnotesize,
    breakatwhitespace=false,         
    breaklines=true,                 
    captionpos=b,                    
    keepspaces=true,                 
    numbers=left,                    
    numbersep=5pt,                  
    showspaces=false,                
    showstringspaces=false,
    showtabs=false,                  
    tabsize=2
}

\allowdisplaybreaks

\newcommand{\eat}[1]{}
\usepackage[normalem]{ulem}
\title{Let the CAT out of the bag:\\ \textbf{C}ontrastive \textbf{A}ttributed explanations for \textbf{T}ext   \thanks{ In the Proceedings of EMNLP 2022.}}

\author{
 Saneem  A. Chemmengath\thanks{$^1$Work done while at IBM Research. } \\
   Microsoft, Bangalore\\
      \texttt{schemmengath@microsoft.com}\\
      \and
   Amar Prakash Azad\\
   IBM Research, Bangalore\\
   \texttt{amarazad@in.ibm.com}
   \and
   Ronny Luss\\
   IBM Research, NY\\
   \texttt{rluss@us.ibm.com}
   \and
   Amit Dhurandhar\\
   IBM Research, NY\\
   \texttt{adhuran@us.ibm.com} }

\date{}
\begin{document}

\maketitle

\begin{abstract}
Contrastive explanations for understanding the behavior of black box models has gained a lot of attention recently as they provide potential for recourse. In this paper, we propose a method \textbf{C}ontrastive \textbf{A}ttributed explanations for \textbf{T}ext (CAT) which provides contrastive explanations for natural language text data with a novel twist as we build and exploit attribute classifiers leading to more semantically meaningful explanations. To ensure that our contrastive generated text has the fewest possible edits with respect to the original text, while also being fluent and close to a human generated contrastive,
we resort to a minimal perturbation approach regularized using a BERT language model and attribute classifiers trained on available attributes. We show through qualitative examples and a user study that our method not only conveys more insight because of these attributes, but also leads to better quality (contrastive) text. 
\eat{Moreover, quantitatively we show that our method is more efficient than other state-of-the-art methods with it also scoring higher on benchmark metrics such as flip rate, (normalized) Levenstein distance, fluency and content preservation.} Quantitatively, we show that our method outperforms other state-of-the-art methods across four data sets on four benchmark metrics.
\end{abstract}

\section{Introduction}
\label{intro}

\begin{table*}[t]
\centering
\small
\caption{Contrastive explanations are shown for state-of-the-art methods GYC and MICE along with our method CAT to explain predictions on two sentences from the AGNews dataset. \textcolor{red}{Red} highlighting indicates text that has changed in the contrastive explanation. Both inputs were classified as Sci-Tech while all contrastive sentences were classified as Business. In addition to these changes, CAT outputs the attributes (or subtopics) that were added/removed from the input to create the contrast. In the first example, ``tax" could correspond to ``Entertainment" or ``Politics" which are attributes added, while ``file" corresponds to software files that are often encrypted and hence to ``Cryptography" (denoted Crypt.) which is removed. In the second example it is easy to see that the added word ``Healthcare" relates to "Medicine" which is added while the removed word ``Search" is related to "Windows" (denoted Wndws) and "Cryptography" which are removed attributes.\eat{  As can be seen, CAT provides much closer and meaningful contrasts, where the attribute information provides further insight into what subtopics were added/removed to produce the contrast. This can unequivocally help appropriate trust in a model.}}
\label{tab:introeg}
\begin{tabular}{|p{2.7cm}|p{3.0cm}|p{2.7cm}|p{2.8cm}|p{2.5cm}|}
  \hline
  \multicolumn{1}{|c|}{\multirow{2}{*}{\textbf{Input}}} & \multicolumn{1}{c|}{\multirow{2}{*}{\textbf{GYC}}} & \multicolumn{1}{c|}{\multirow{2}{*}{\textbf{MICE}}} & \multicolumn{2}{c|}{\textbf{CAT}} \\ \cline{4-5}
& & &\multicolumn{1}{c|}{Contrast} & \multicolumn{1}{c|}{Attributes} \\
 \hline
\eat{  Global server sales on the rise & Global server sales on \textcolor{red}{Friday, October} & Global \textcolor{red}{DS} sales on the rise & Global \textcolor{red}{ticket} sales on the rise & $+$Travel, $+$Sale, $-$Computer &Global \textcolor{red}{drug} sales on the rise \\
\hline}
\eat{ U S Justice Department Cracks Down Internet Crime & U S Justice Department Cracks Down \textcolor{red}{ and in,} & U S Justice Department Cracks Down \textcolor{red}{Business} Crime & U S Justice Department Cracks Down \textcolor{red}{Economic} Crime & $+$Politics  \\}
  Movie Studios to sue illegal film file traders & Movie Studios to sue \textcolor{red}{juveniles psychiatrically}& Movie Studios to sue illegal film \textcolor{red}{- Investors} & Movie Studios to sue illegal film \textcolor{red}{tax} traders &
  $+$Entertainment, $+$Politics, $-$Crypt. \\
  \hline
  Search providers seek video find challenges & Search providers seek video find \textcolor{red}{videos,} & seek \textcolor{red}{opportunities} find challenges & \textcolor{red}{Healthcare} providers seek \textcolor{red}{to} find challenges & $+$Medicine, $-$Wndws, $-$Crypt.  \\
 \hline
 \end{tabular}
 \end{table*}

Explainable AI (XAI) has seen an explosion of interest over the last five years, not just in research \cite{molnarbook,arya2019explanation}, but also in the real world where governments \cite{gdpr,xai} and industry have made sizeable investments. The primary driver for this level of interest has been the inculcation of deep learning technologies \cite{gan}, which are inherently black box, into decision making systems that affect billions of people. Trust thus has become a central theme in relying on these black box systems, and one way to achieve it is seemingly through obtaining explanations.

Although many feature-based \cite{lime,unifiedPI,saliency} and exemplar-based methods \cite{proto,infl,l2c} have been proposed to explain local instance level decisions of black box models, contrastive/counterfactual explanations have seen a surge of interest recently \cite{gdpr-wachter,CEM,gyc,CEM-MAF,mice}. One reason for this is that
contrastive explanations are viewed as one of the main tools to achieve recourse \cite{algorec}. For example, companies commonly use chatbots to communicate with customers, which starts by passing customer text through a classifier that decides which support department should handle the customer. A common problem is locating bias in these classifier models since ``the chatbot will continue to show the behavior” due to biased knowledge bases\cite{chatbot_bias}; recourse in terms of removing bias could be achieved by identifying examples that drive the bias.

Given this surge of interest and its importance in recourse, in this paper, we propose a novel method \textbf{C}ontrastive \textbf{A}ttributed explanations for \textbf{T}ext (CAT) which provides contrastive explanations for natural language data, a modality that has received comparatively less attention when it comes to these type of explanations. We show that our method produces fluent contrasts and possesses an additional novel twist not seen in prior works. As such, our method also outputs a minimal set of semantically meaningful attributes that it thinks led to the final contrast. These attributes could be subtopics in a dataset, different from the class labels, that characterize a piece of text or the attributes could even be obtained from a different dataset. Our approach is to leverage these attributes by building models (viz. classifiers) for them and then using these classifiers to guide the search for contrasts in the original task. Regarding the motivating biased chatbot example above, learning attributes that lead to contrasts can be particularly useful. Gender
bias, for instance, is hard to locate because contrasts
can modify gender on individual chats in different ways (removing pronouns, replacing gendered words with neutral, etc.). Such bias can be more easily identified if the different modifications all affect a (latent) attribute that is gendered such as motherhood, which our CAT explanations should highlight.

To better understand what CAT offers, consider the examples provided in Table \ref{tab:introeg}. Here we show two example sentences from the AG News dataset \cite{AgNews2015} which were classified as Sci-Tech by our neural network black box (BB) model (details in Section \ref{sec:exp}). Each explanation method generates a contrast in the class Business where the other choices were World and Sports. As can be seen, our method CAT produces closer contrasts than two recent methods: Generate Your Counterfactuals (GYC) \cite{gyc} and Minimal Contrastive Editing (MICE) \cite{mice}. 

A key novelty is that CAT provides additional information in terms of characteristic attributes it thinks the contrast sentence belongs to (indicated by $+$ sign), while also indicating characteristics it thinks are no longer applicable (indicated by $-$ sign).\eat{ These attributes were obtained from the 20 Newsgroup dataset \cite{AgNews2015} and the Huffpost News-Category dataset \cite{news_category}, where we trained classifiers for 42 such attributes (20 Newsgroup classes $+$ 22 Huffpost classes), and used them to regularize our search for the contrastive text.} Our method picks a few relevant attributes that guide generation of the contrast; the attributes themselves provide additional insight into the functioning of the black box model. This is confirmed through two separate user studies conducted with a total of 75 participants for which the results are reported in section \ref{sec:human}. Users found it easier to predict the class of the input sentence given our explanation over GYC, MICE, MICE with no fine tuning (MICE-nft), and an ablation of our method, called CAT with no attributes (CAT-na), and moreover, users qualitatively preferred our method in terms of understandability, sufficiency, satisfiability and completeness on a five point Likert scale.

It is important to note that there are various ways to generate text using different language models (GPT-2, BERT, T5, etc.) and even different techniques on how infilling might be performed (forward or bi-directional). The key idea of guiding the generation of contrasts through attributes can be considered for other recent methods, whether the contrast is learned through differentiable optimization \cite{gyc} or through combinatorial search \cite{mice}. 

\eat{The general applicability of CAT is also important to note relative to GYC and MICE.} We note that author provided implementations of GYC and MICE require models that use specific text embeddings and are not easily adaptable to other embeddings. Our method CAT, however, is easily adaptable, and we thus compare against the methods GYC and MICE using the embeddings they are respectively implemented for in order to get stringent and fair comparisons for CAT with previous state-of-the-art methods. This means we must compare against GYC and MICE on different text classification models, and hence require two separate user studies, one comparing CAT with GYC and the other comparing CAT with MICE (and we suspect this is why no comparisons are found in the literature). 

As such, our contributions are as follows:
    1) CAT introduces the idea of using attributes to drive the generation of text contrastives. This contribution is both conceptual as it brings new insight to the user as well as methodological as it leads to user-preferred contrasts as seen in the user study.
    2) The CAT implementation is easily adaptable to classifiers with different embeddings.
    3) We qualitatively evaluate CAT through examples on four different datasets from different domain tasks.
    4) We quantitatively evaluate CAT over other methods in terms of flip rate, content preservation, fluency, Levenstein distance and efficiency.
    5) We demonstrate the utility of CAT through two user studies that ask users to determine a model's prediction on an instance given an explanation; CAT is compared with GYC, MICE, MICE with no fine tuning, and an ablation of our method CAT-na.




\eat{
\begin{table*}[htbp]
\centering
\caption{Below we see contrastive explanations for a state-of-the-art method GYC, our method CAT and an ablation of our method without using attributes CAT-na for two sentences from the AGNews dataset. The \textcolor{red}{red} highlighting indicates the text that has changed in the contrastive explanation. The inputs were classified as Sci-Tech, while the contrasts for the different methods are all classified as Business. For CAT besides these changes we also see the attributes (or subtopics) that it thinks were added/removed from the input to create the contrast. For instance, in the first example it is easy to to see that the added word ``ticket" could relate to ``Travel" and ``Sale", while the removed word ``server" from the input relates to ``Computers". Similarly, ``tax" in the second example could correspond to ``Entertainment" or ``Politics", while ``file" corresponds to software files which many times are encryted and hence to ``Cryptography".  As can be seen CAT provides much closer and meaningful contrasts, where the attribute information provides further insight into what subtopics were added/removed to produce the contrast. This can unequivocally help appropriate trust in a model.}
\label{tab:introeg}
\begin{tabular}{|p{4cm}|p{4cm}|p{4cm}|p{4cm}|}
  \hline
   \textbf{Input} & \textbf{GYC} & \textbf{CAT} & \textbf{CAT-na} \\
 \hline
  Global server sales on the rise & Global server sales on \textcolor{red}{Friday, October} & Global \textcolor{red}{ticket} sales on the rise & Global \textcolor{red}{drug} sales on the rise \\
  &&&\\
 && Attributes: $+$Travel, $+$Sale, $-$Computer &\\
 \cline{1-4}
  Movie Studios to sue illegal film file traders & Movie Studios to sue \textcolor{red}{juveniles psychiatrically} & Movie Studios to sue illegal film \textcolor{red}{tax} traders &
  Movie studios to sue illegal \textcolor{red}{insurance} file traders\\
 &&&\\
 && Attributes: $+$Entertainment, $+$Politics, $-$Cryptography&\\
 \hline
 \end{tabular}
 \end{table*}
 }





\begin{figure}[!tb]
\centering
\includegraphics[width=.95\textwidth]{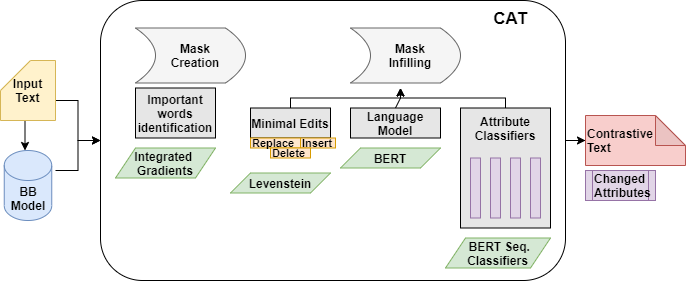}
\caption{Above we see how CAT creates contrastive explanations. It first finds important words in the text which it then (minimally) alters by either replacing/deleting a subset of them or by inserting new ones in the spaces around them so as to i) change the black box model's prediction, ii)  maintain fluency and iii) change a minimal number of attribute classifier predictions (by a significant amount). The output is the contrastive text along with a list of the attributes added/removed relative to the input text.}
  \label{fig:cat}
\end{figure}

\section{Related Literature}
\label{s:rel}

Regarding the explanations of machine learning predictions on text data, a recent survey \cite{xainlp_survey} considered 50 recent explainability works for natural language processing, and moreover only methods that ``justify predictions" rather than understanding ``a model's behavior in general". Our intention is also to explain individual predictions. Little work has been done for global explainability with respect to text classification; \cite{lime} suggests using various ``representative" local predictions to get a global understanding of model behavior. The vast majority of explainability techniques found by \cite{xainlp_survey} fall under local explainability.

Local methods can be divided among post-hoc methods that explain a fixed model's prediction and self-explainable methods where the model is itself understandable; our focus is on the former. One large group of explainability methods are feature based where the explanation outputs some form of feature importance (i.e., ranking, positive/negative relevance, contributions, etc.) of the words in text \cite{dknn_text,dknn,inputreduction,patternattribution_text,lime,melis_jaakkola_2017}. 
Other types of local post-hoc explanations include exemplar based \cite{proto,infl,l2c} that output similar instances to the input.

Amongst local methods our focus is on contrastive/counterfactual methods \cite{CEM, gyc, CEM-MAF} that modify the input such that the class changes and explains the prediction as ``If the sample were modified by $X$, the prediction would have been $Y$ instead," where $X$ is a change to the input and $Y$ is a new class. Such explanations are complementary to the other methods discussed above. They offer different types of intuition and thus should be used in conjunction with rather than instead of counterfactuals. \cite{CEM-MAF} learn contrasts using latent attributes, but specifically for color images. Their approach does not readily translate to text as humans do not perceive minor unrealistic parts of a generated image, whereas any nuance in text is easily noticed. Thus, generating (fluent) contrasts for text models is inherently much more challenging.

The most relevant comparisons to our work are GYC \cite{gyc} and MICE \cite{mice}. GYC builds on the text generation of \cite{pplm} by learning perturbations to the history matrix used for language modeling that are also trained to reconstruct input from the generated counterfactuals. A diversity regularization term ensures that this does not result in counterfactuals that are identical to the input. A more recent work MICE masks and replaces important words selected by \cite{saliency} in order to flip the prediction. MICE requires fine-tuning their language model to each dataset, which is a significant overhead especially given the fact that we are simply generating local explanations versus our CAT framework. Not to mention in many real applications (sufficient) data may not be available to fine tune explanations \cite{macem}. Both GYC and MICE works are targeted, i.e., the user must decide what class the counterfactual should be in as opposed to CAT which automatically decides the contrast class. 

Other recent methods are POLYJUICE \cite{polyjuice} and a contrastive latent space method \cite{jacovi}. The former is a human-in-the-loop method requiring supervision about the type of modification to be performed to the text such as negation, word replacement, insertion, deletion, and is not catered towards explaining a specific classifier by automatically finding the appropriate edits. The latter does not generate contrastive text but rather highlights (multiple) words in the input text that are most likely to alter the prediction if changed, where again the target class has to be provided. Further, \cite{jacovi} assume access to encodings from the second-to-last layer of the model being explained, and is thus not a black box method like CAT. 
Our focus being automated contrastive explanation generation, we compare with GYC and MICE, i.e., methods designed towards explaining a classifier, where a valid contrast is also generated. The value of CAT versus GYC and MICE comes from the output of (hidden) subtopics that are added/removed from the original text to create the contrast, giving important intuition that is missing from these other methods. This is confirmed through two user studies we conduct and qualitative examples we provide. The attribute classifiers built to predict these subtopics also aid in creating better contrasts. Moreover, as will be evident, such attribute classifiers can be used across datasets, thus precluding the need for each dataset to contain such attributes. Furthermore, topic models or autoencoders could be used to divulge such attributes.

\eat{
\begin{table*}[htbp]
\centering
\small
\caption{Five examples of CAT applied to AGNews dataset. Modifications to/in original/contrasts are shown by green/red highlighting. Attribute changes offer insight as to how the black box model views certain words with multiple possible meanings.}
\label{tab:agnews}
\begin{tabular}{|c|c|c|c|c|}
\hline
 \multirow{2}{*}{\textbf{Input}}& \multirow{2}{*}{\textbf{CAT}} & \textbf{Attribute}  & \textbf{Input} & \textbf{Contrast} \\
 &  & \textbf{Changes} & \textbf{Pred}  & \textbf{Pred} \\ \hline

\eat{\multirow{3}{*}{\shortstack{New Hummer Is Smaller, Cheaper\\and Less \textcolor{green}{Gas} Hungry}} & \multirow{3}{*}{\shortstack{New Hummer Is Smaller, Cheaper\\and Less \textcolor{red}{Tax} Hungry}} & \multirow{3}{*}{\shortstack{+cryptography\\-travel\\-space}} & \multirow{3}{*}{business} & \multirow{3}{*}{sci-tech} \\
& & & & \\
& & & & \\ \hline}

\eat{\multirow{2}{*}{\shortstack{\textcolor{green}{Oil} Up from 7-Week Lows on U.S.\\ Weather}} & \multirow{2}{*}{\shortstack{\textcolor{red}{Winds} Up from 7-Week Lows on U.S.\\ Weather}} & \multirow{2}{*}{+travel} & \multirow{2}{*}{business} & \multirow{2}{*}{world} \\
& & & & \\ \hline}

\multirow{6}{*}{\shortstack{Kazaa Owner Cheers \textcolor{green}{File}-Swapping\\ Decision (AP)}}& \multirow{6}{*}{\shortstack{Kazaa Owner Cheers \textcolor{red}{Salary}-Swapping\\ Decision (AP)}} & \multirow{6}{*}{\shortstack{+forsale\\+baseball\\+hockey\\-arts\\-windows\\-cryptography}} & \multirow{6}{*}{sci-tech} & \multirow{6}{*}{sports} \\
& & & & \\
& & & & \\
& & & & \\
& & & & \\
& & & & \\ \hline

\multirow{2}{*}{New Human Species \textcolor{green}{Discovered}} & \multirow{2}{*}{New Human Species \textcolor{red}{influenza}} & \multirow{2}{*}{\shortstack{+medicine\\-cryptography}} & \multirow{2}{*}{sci-tech} & \multirow{2}{*}{world} \\
& & & & \\ \hline

\eat{\multirow{5}{*}{\shortstack{Perfect start for France in \textcolor{green}{Federation}\\ Cup}} & \multirow{5}{*}{\shortstack{Perfect start for France in \textcolor{red}{microsoft}\\ Cup}} & \multirow{5}{*}{\shortstack{+cryptography\\-motorcycles\\-hockey\\-space\\-politics}} & \multirow{5}{*}{sports} & \multirow{5}{*}{sci-tech} \\
& & & & \\
& & & & \\
& & & & \\
& & & & \\ \hline}

\eat{\multirow{3}{*}{\shortstack{\textcolor{green}{Kuwait}: Fundamentalists Recruiting \\ Teens (AP)}} & \multirow{3}{*}{\shortstack{\textcolor{green}{Source}: Fundamentalists Recruiting \\ Teens (AP)}} & \multirow{3}{*}{\shortstack{+guns\\+religion\\-mideast}} & \multirow{3}{*}{world} & \multirow{3}{*}{sci-tech} \\
& & & & \\
& & & & \\ \hline}

\eat{\multirow{4}{*}{\shortstack{Bribes cited in downing of \\ \textcolor{green}{Russian} jets}} & \multirow{4}{*}{\shortstack{Bribes cited in downing of \\ \textcolor{red}{houston} jets}} & \multirow{4}{*}{\shortstack{+cryptography\\+politics\\-arts\\-space}} & \multirow{4}{*}{world} & \multirow{4}{*}{business} \\
& & & & \\
& & & & \\
& & & & \\ \hline}

\multirow{3}{*}{\shortstack{US shows flexibility on \textcolor{green}{Israeli}\\ settlements}} & \multirow{3}{*}{\shortstack{US shows flexibility on \textcolor{red}{virtual}\\ settlements}} & \multirow{3}{*}{\shortstack{+arts\\+cryptography\\-mideast}} & \multirow{3}{*}{world} & \multirow{3}{*}{sci-tech} \\
& & & & \\
& & & & \\ \hline

\eat{\multirow{3}{*}{Harry \textcolor{green}{in nightclub} scuffle} & 
\multirow{3}{*}{Harry \textcolor{red}{inflight} scuffle} & 
\multirow{3}{*}{\shortstack{+sports\\+space\\-cryptography}} & \multirow{3}{*}{world} & \multirow{3}{*}{sci-tech} \\
& & & & \\
& & & & \\ \hline}

\multirow{2}{*}{\shortstack{Will sinking Lowe resurface in\\ \textcolor{green}{playoffs}?}} & \multirow{2}{*}{\shortstack{Will sinking Lowe resurface in\\ \textcolor{red}{2020}?}} & \multirow{2}{*}{\shortstack{+space\\-hockey}} & \multirow{2}{*}{sports} & \multirow{2}{*}{business} \\
& & & & \\ \hline

\multirow{3}{*}{\shortstack{Pace of U.S. \textcolor{green}{Factory} Growth Climbs\\ in Dec}} & \multirow{3}{*}{\shortstack{Pace of U.S. \textcolor{red}{population} Growth\\ Climbs in Dec}} & \multirow{3}{*}{\shortstack{+politics\\-money\\-travel}} & \multirow{3}{*}{business} & \multirow{3}{*}{sci-tech} \\
& & & & \\
& & & & \\ \hline

\end{tabular}
 \end{table*}
}

\eat{
\begin{table*}[t]
\small
\centering
\caption{Five examples of CAT applied to Hate Speech dataset. Modifications to/in original/contrasts are shown by strikeout/red highlighting. \$\$\$\$ represents a vulgar word.\eat{ Anonymous links and twitter handles were removed.} No change to Input means there was an insertion.}
\label{tab:hate}
\begin{tabular}{|c|c|c|c|c|}
\hline
 \multirow{2}{*}{\textbf{Input}}& \multirow{2}{*}{\textbf{CAT}} & \textbf{Attribute}  & \textbf{Input} & \textbf{Contrast} \\
 &  & \textbf{Changes} & \textbf{Pred}  & \textbf{Pred} \\ \hline

Ever try to make a \textcolor{green}{girl} play video &Ever try to make a \textcolor{red}{patient} play video &\multirow{4}{*}{\shortstack{+medicine \\ -parenting \\ -arts}} &\multirow{4}{*}{hate} &\multirow{4}{*}{not hate} \\
games? they are terrible. these are the &games? they are terrible. these are the && & \\
skills they will need for the future.&skills they will need for the future. & & & \\
we're \$\$\$\$.@ &we're \$\$\$\$.@ & & & \\ \hline

\eat{
Brazil-Venezuela border reopens amid&Brazil-Venezuela border reopens amid & & &\\
fight over refugees https://t.co/ &fight over \textcolor{red}{illegal} refugees https://t.co/  & & & \\
rKitw9zoDirKitw9zoDi \#FoxNews @vp &rKitw9zoDirKitw9zoDi \#FoxNews @vp  & & & \\ \hline
}

It sounds so ugly to hear a \textcolor{green}{female} &It sounds so ugly to hear a \textcolor{red}{parent} &+parenting &\multirow{4}{*}{hate}  & \multirow{4}{*}{not hate} \\
talk about another female claiming & talk about another female claiming &+atheism & & \\
she’s a “\$\$\$\$ or \$\$\$\$” like \$\$\$\$ &she’s a “\$\$\$\$ or \$\$\$\$” like \$\$\$\$  &-entertainment & &  \\
it ain’t yo \$\$\$\$ why u worried for &it ain’t yo \$\$\$\$ why u worried for  &-politics & &  \\ \hline

\eat{
ILLEGAL ALIEN CRIMEIs rampant in the& & & & \\
first Sanctuary State 'Oregon'!These statistic & & & & \\
are outrageous and should be construed & & & & \\
to be a close poll of US states.@SpeakerRyan & & & & \\
\#NoAmnesty \#NoDACA \#DeportThemAll \#ProtectUSCitizens & & & &
}

Illegal \textcolor{green}{immigration} drives up real &Illegal \textcolor{red}{software} drives up real &\multirow{6}{*}{\shortstack{+money \\ +science}}  &\multirow{6}{*}{hate}  &\multirow{6}{*}{not hate}  \\
estate prices, widens wealth-gaps,&estate prices, widens wealth-gaps,  & & & \\
reduces high-tech investment, increases&reduces high-tech investment, increases& & & \\
 state and local tax burdens, hurts kids'& state and local tax burdens, hurts kids'& & & \\
schools and college education, pushes&schools and college education, pushes&& & \\
Americans away from high-tech jobs&Americans away from high-tech jobs& & &\\ \hline

\multirow{3}{*}{\shortstack{When \textcolor{green}{you}'re out drinking all night\\vs when the bill comes }} & 
\multirow{3}{*}{\shortstack{When \textcolor{red}{women}'re out drinking all night\\vs when the bill comes }}& +politics&\multirow{3}{*}{not hate}  &\multirow{3}{*}{hate}  \\
& &-environment && \\
& &-guns & &\\ \hline

I thought you were leaving the&I thought you \textcolor{red}{ladies} were leaving the &\multirow{2}{*}{+politics} &\multirow{2}{*}{not hate} &\multirow{2}{*}{hate} \\
country you washed up \$\$\$\$ \$\$\$\$? &country you washed up \$\$\$\$ \$\$\$\$? & & & \\\hline
\end{tabular}
\end{table*}
}

\eat{\begin{table*}[t]
\small
\centering
\caption{Five examples of CAT applied to Hate Speech dataset. Modifications to/in original/contrasts are shown by strikeout/red highlighting. \$\$ represents a vulgar word.\eat{ Anonymous links and twitter handles were removed.} No strikeout implies there was an insertion.}
\label{tab:hate}
\begin{tabular}{|p{8.5cm}|c|p{3cm}|p{3cm}|}
\hline
\multicolumn{1}{|c|}{\multirow{2}{*}{\textbf{Input/CAT}}} & \multicolumn{1}{c|}{\multirow{2}{*}{\textbf{Attribute Changes}}}  & \multicolumn{1}{c|}{\textbf{Input}} & \multicolumn{1}{c|}{\textbf{Contrast}} \\
   &  & \multicolumn{1}{c|}{\textbf{Pred}}  &\multicolumn{1}{c|}{\textbf{Pred}} \\ \hline

Ever try to make a \st{girl} \textcolor{red}{patient} play video games? they are terrible. these are the skills they will need for the future. we're \$\$.@ & \multirow{2}{*}{\shortstack{+medicine, -arts,\\  -parenting}} & \multicolumn{1}{c|}{hate} & \multicolumn{1}{c|}{not hate}\\ \hline

It sounds so ugly to hear a \st{female} \textcolor{red}{parent} talk about another female claiming she’s a “\$\$ or \$\$” like \$\$ it ain’t yo \$\$ why u worried for& \multirow{2}{*}{\shortstack{+parenting, +atheism, \\-entertainment, -politics}} &\multicolumn{1}{c|}{\multirow{2}{*}{hate}}  & \multicolumn{1}{c|}{\multirow{2}{*}{not hate}} \\ \hline

Illegal \st{immigration} \textcolor{red}{software} drives up real estate prices, widens wealth-gaps, reduces high-tech investment, increases state and local tax burdens, hurts kids' schools and college education, pushes Americans away from high-tech jobs & \multirow{4}{*}{+money, +science} &\multicolumn{1}{c|}{\multirow{4}{*}{hate}}  & \multicolumn{1}{c|}{\multirow{4}{*}{not hate}} \\ \hline

When \st{you} \textcolor{red}{women}'re out drinking all night vs when the bill comes & +politics, -environ., -guns &\multicolumn{1}{c|}{not hate}  & \multicolumn{1}{c|}{hate}  \\
 \hline

\eat{I thought you \textcolor{red}{ladies} were leaving the country you washed up \$\$\$\$ \$\$\$\$&+politics &\multicolumn{1}{c|}{not hate}  & \multicolumn{1}{c|}{hate}  \\ \hline}

\multirow{2}{*}{Readers respond: Kudos on \textcolor{red}{illegal} immigrant \st{features} \textcolor{red}{women} \quad\quad\quad\quad\quad\quad} &\multirow{2}{*}{\shortstack{+latino voices, +politics\\-travel, -mideast}} &\multicolumn{1}{c|}{\multirow{2}{*}{not hate}} & \multicolumn{1}{c|}{\multirow{2}{*}{hate}}  \\ 
\multicolumn{1}{|c|}{}&\multicolumn{1}{c|}{}&\multicolumn{1}{c|}{}&\multicolumn{1}{c|}{}\\ \hline 
\end{tabular}
\end{table*}
}
\eat{
\begin{table*}[htbp]
\centering
\small
\caption{Five examples of CAT applied to NLI dataset. Modifications to/in original/contrasts are shown by green/red highlighting. The $<$/s$>$ delimiter separates text from hypothesis. Attribute changes offer insight as to how the black box model views certain words that could have multiple meanings.}
\label{tab:nli}
\begin{tabular}{|c|c|c|c|c|}
\hline
 \multirow{2}{*}{\textbf{Input}}& \multirow{2}{*}{\textbf{CAT}} & \textbf{Attribute}  & \textbf{Input} & \textbf{Contrast} \\
 &  & \textbf{Changes} & \textbf{Pred}  & \textbf{Pred} \\ \hline

Two outdoor workers conversing while& Two outdoor workers conversing while &\multirow{2}{*}{-electronics} &\multirow{2}{*}{\shortstack{entail\\-ment}} & \multirow{2}{*}{\shortstack{contra\\-diction}} \\ 
on \textcolor{green}{break}.$<$/s$>$ people on sidewalk &on \textcolor{red}{lawn}.$<$/s$>$people on sidewalk &  &  &\\ \hline

The double sink is freshly polished&The double sink is freshly polished&+space &\multirow{3}{*}{\shortstack{entail\\-ment}} & \multirow{3}{*}{\shortstack{contra\\-diction}}\\
chrome.$<$/s$>$ The sink is \textcolor{green}{chrome} &chrome. $<$/s$>$ The sink is \textcolor{red}{rust} &-graphics & & \\
colored.&colored.&-electronics & & \\ \hline

A group of children, wearing white&A group of children, wearing white&\multirow{4}{*}{-entertainment} &\multirow{4}{*}{neutral} & \multirow{4}{*}{\shortstack{contra \\ -diction}}\\
karate shirts, look at the American flag.&karate shirts, look at the American flag.& & & \\
$<$/s$>$The children \textcolor{green}{are} at a karate&$<$/s$>$The children \textcolor{red}{looked} at a karate& & & \\
tournament. &tournament & & & \\ \hline

A man is about to \textcolor{green}{play} his guitar.$<$/s$>$&A man is about to \textcolor{red}{steal} his guitar.$<$/s$>$& \multirow{2}{*}{-entertainment}&\multirow{2}{*}{neutral} &contra \\
a man is performing for school children&a man is performing for school children&&&-diction \\ \hline

Two women are giving each other a&Two women are giving each other a&\multirow{5}{*}{\shortstack{+graphics \\ -electronics}} &\multirow{5}{*}{\shortstack{contra \\ -diction}} &\multirow{5}{*}{neutral} \\
hug while a man holding a glass is&hug while a man holding a glass is& & & \\
looking at the camera.$<$/s$>$ The &looking at the camera.$<$/s$>$ The& & & \\
people are all taking \textcolor{green}{naps} during the&people are all taking \textcolor{red}{photos} during&&& \\
hottest part of the day. & the hottest part of the day. &&& \\
\hline
\end{tabular}

\end{table*}
}
\eat{
\begin{table*}[htbp]
\centering
\small
\caption{Five examples of CAT applied to NLI dataset. Modifications to/in original/contrasts are shown by green/red highlighting. The $<$/s$>$ delimiter separates text from hypothesis. Attribute changes offer insight as to how the black box model views certain words that could have multiple meanings.}
\label{tab:nli}
\begin{tabular}{|c|c|c|c|c|}
\hline
 \multirow{2}{*}{\textbf{Input}}& \multirow{2}{*}{\textbf{CAT}} & \textbf{Attribute}  & \textbf{Input} & \textbf{Contrast} \\
 &  & \textbf{Changes} & \textbf{Pred}  & \textbf{Pred} \\ \hline

Two outdoor workers conversing while& Two outdoor workers conversing while &\multirow{2}{*}{-electronics} &\multirow{2}{*}{\shortstack{entail\\-ment}} & \multirow{2}{*}{\shortstack{contra\\-diction}} \\ 
on \textcolor{green}{break}.$<$/s$>$ people on sidewalk &on \textcolor{red}{lawn}.$<$/s$>$people on sidewalk &  &  &\\ \hline

The double sink is freshly polished&The double sink is freshly polished&+space &\multirow{3}{*}{\shortstack{entail\\-ment}} & \multirow{3}{*}{\shortstack{contra\\-diction}}\\
chrome.$<$/s$>$ The sink is \textcolor{green}{chrome} &chrome. $<$/s$>$ The sink is \textcolor{red}{rust} &-graphics & & \\
colored.&colored.&-electronics & & \\ \hline

A group of children, wearing white&A group of children, wearing white&\multirow{4}{*}{-entertainment} &\multirow{4}{*}{neutral} & \multirow{4}{*}{\shortstack{contra \\ -diction}}\\
karate shirts, look at the American flag.&karate shirts, look at the American flag.& & & \\
$<$/s$>$The children \textcolor{green}{are} at a karate&$<$/s$>$The children \textcolor{red}{looked} at a karate& & & \\
tournament. &tournament & & & \\ \hline

A man is about to \textcolor{green}{play} his guitar.$<$/s$>$&A man is about to \textcolor{red}{steal} his guitar.$<$/s$>$& \multirow{2}{*}{-entertainment}&\multirow{2}{*}{neutral} &contra \\
a man is performing for school children&a man is performing for school children&&&-diction \\ \hline

Two women are giving each other a&Two women are giving each other a&\multirow{5}{*}{\shortstack{+graphics \\ -electronics}} &\multirow{5}{*}{\shortstack{contra \\ -diction}} &\multirow{5}{*}{neutral} \\
hug while a man holding a glass is&hug while a man holding a glass is& & & \\
looking at the camera.$<$/s$>$ The &looking at the camera.$<$/s$>$ The& & & \\
people are all taking \textcolor{green}{naps} during the&people are all taking \textcolor{red}{photos} during&&& \\
hottest part of the day. & the hottest part of the day. &&& \\

\hline
\end{tabular}

\end{table*}
}
\eat{
\begin{table*}[htbp]
\centering
\small
\caption{Five examples of CAT applied to NLI dataset. Modifications to/in original/contrasts are shown by strikeout/red highlighting. The $<$/s$>$ delimiter separates text from hypothesis. Attribute changes offer insight into how the black box model views certain words that could have multiple meanings.}
\label{tab:nli}
\begin{tabular}{|p{8.5cm}|p{3cm}|p{3cm}|p{3cm}|}
\hline
\multicolumn{1}{|c|}{\multirow{2}{*}{\textbf{Input/CAT}}} & \multicolumn{1}{c|}{\multirow{2}{*}{\textbf{Attribute Changes}}}  & \multicolumn{1}{c|}{\textbf{Input}} & \multicolumn{1}{c|}{\textbf{Contrast}} \\
   &  & \multicolumn{1}{c|}{\textbf{Pred}}  &\multicolumn{1}{c|}{\textbf{Pred}} \\ \hline

Two outdoor workers conversing while on \st{break} \textcolor{red}{lawn}. $<$/s$>$ people on sidewalk &\multicolumn{1}{c|}{\multirow{2}{*}{-electronics}} &\multicolumn{1}{c|}{\multirow{2}{*}{\shortstack{entail\\-ment}}} & \multicolumn{1}{c|}{\multirow{2}{*}{\shortstack{contra\\-diction}}} \\ \hline

The double sink is freshly polished chrome. $<$/s$>$ The sink is \st{chrome} \textcolor{red}{rust} colored&\multicolumn{1}{c|}{\multirow{2}{*}{\shortstack{+space, -graphics \\-electronics}}}
 &\multicolumn{1}{c|}{\multirow{2}{*}{\shortstack{entail\\-ment}}} & \multicolumn{1}{c|}{\multirow{2}{*}{\shortstack{contra\\-diction}}}\\ \hline

A group of children, wearing white karate shirts, look at the American flag.$<$/s$>$ The children \st{are} \textcolor{red}{looked} at a karate tournament.&\multicolumn{1}{c|}{\multirow{2}{*}{-entertainment}} &\multicolumn{1}{c|}{\multirow{2}{*}{neutral}} & \multicolumn{1}{c|}{\multirow{2}{*}{\shortstack{contra \\ -diction}}} \\ \hline

A man is about to \st{play} \textcolor{red}{steal} his guitar.$<$/s$>$ a man is performing for school children& \multicolumn{1}{c|}{\multirow{2}{*}{-entertainment}}&\multicolumn{1}{c|}{\multirow{2}{*}{neutral}} & \multicolumn{1}{c|}{\multirow{2}{*}{\shortstack{contra \\ -diction}}} \\ \hline

Two women are giving each other a hug while a man holding a glass is looking at the camera. $<$/s$>$ The people are all taking \st{naps}  \textcolor{red}{photos} during the hottest part of the day. & \multicolumn{1}{c|}{\multirow{3}{*}{+graphics, -electronics}}& \multicolumn{1}{c|}{\multirow{3}{*}{\shortstack{contra \\ -diction}}} & \multicolumn{1}{c|}{\multirow{3}{*}{neutral}} 
\\ \hline
\end{tabular}

\end{table*}
}


\section{Proposed Approach}
\label{s:approach}
We now describe our \textbf{C}ontrastive \textbf{A}ttributed explanations for \textbf{T}ext  (\textbf{CAT}) method. Contrastive explanations, convey why the model classified a certain input instance to a class $p$, and not another class $q$. This is achieved by creating contrastive examples (also called contrasts) from input instance which get predicted as $q$. Contrastive examples are created by minimally perturbing the input such that the model prediction changes. In the case of text data, perturbations can be of three types: (1) inserting a new word, (2) replacing a word with another, and (3) deleting a word. In addition to keeping the number of such perturbations small, constrastive explainers also try to maintain grammatical correctness and fluency of the contrasts \cite{gyc, mice}. 

As an example, take the case of a black box model trained on the AG News dataset that predicts which category a certain news headline falls under. Given a headline, 
``\textit{Many technologies may be a waste of time and money, researcher says}" 
which is predicted as Sci-Tech, a contrastive explainer will try to explain why this headline wasn't predicted as, say, \textit{Business} by generating a contrastive example,
``\textit{Many \st{technologies} \textcolor{red}{jobs} may be a waste of time and money, researcher says}" 
which is predicted as Business. Observe that a single word replacement achieves a prediction change. Such contrastive explanations can help users test the robustness of black box classification models.

We observed that even with constraints for minimal perturbation and fluency on a given black box model and an instance, there are multiple contrastive examples to choose from and, very often, many are less informative than others. For example, another possible contrast is, 
``\textit{Many technologies may be a waste of \st{time} \textcolor{red}{investment} and money , researcher says"} 
which also gets predicted as Business. However, this particular explanation is not as intuitive as the previous one as ``money" is a form of ``investment" and the nature of the sentence has not changed in an obvious sense with the word ``technologies" still present in the sentence. 

To alleviate this problem, we propose to construct and use a set of \textit{attribute classifiers}, where the attributes could be tags/subtopics relevant to the classification task 
obtained from the same or a related dataset used to build the original classifier. Attribute classifiers indicate the presence/absence of a certain subtopic in the text and confidence scores from these classifiers could be used as a regularization to create a contrast. We thus prefer contrasts which change attribute scores measurably as opposed to those contrasts which do not. However, at the same time, we want a minimal number of attribute scores to change so as to have crisp explanations. Hence, our regularization not only creates more intuitive contrasts, but also provides additional information to the user in terms of changed subtopics which, as confirmed through our user study in Section \ref{sec:human}, provide better understanding of the model behavior. The important steps in our method are depicted in Figure \ref{fig:cat}.

Formally, given an input text $x \in \mathcal{X}$, and a text classification model $f(\cdot)$ which predicts $y = f(x)\in \mathcal{Y}$, we aim to create a perturbed instance $x'$ such that the predictions $f(x) \neq f(x')$ and $x'$ is ``minimally" different from $x$. We use a set of $m$ attribute classifiers $\zeta_i: \mathcal{X} \rightarrow \mathbb{R} , \forall i \in \{1,\ldots,m\}$, which produce scores indicative of presence (higher scores) or absence (lower scores) of corresponding attributes in the text. We say that attribute $i$ is added to the perturbed sentence if $\zeta_i(x') - \zeta_i(x) > \tau$ and removed when $\zeta_i(x') - \zeta_i(x) < -\tau$, for a fixed $\tau > 0$. Word-level Levenshtein distance between original and perturbed instance $d_{Lev}(x',x)$, which is the minimum number of deletions, substitutions, or insertions required to transform $x$ to $x'$ is used to keep the perturbed instance close to the original. The naturalness (fluency) of a generated sentence $x'$ is quantified by the likelihood of sentence $x'$ as measured by the language model used for generation; we denote  this likelihood by $p_{\text{LM}}(x')$.
For a predicate $\phi$, we denote $\mathbbm{1}_\phi$ the indicator of $\phi$, which takes the value 1 if $\phi$ is true and $0$ otherwise. Given this setup, we propose to find contrastive examples by solving the following optimization problem:
\begin{align}
\label{eqn:main}
    \max_{x' \in \mathcal{X}}&~~~  ||\zeta(x') - \zeta(x)||_{\infty}-\beta\sum_{i} \mathbbm{1}_{|\zeta_i(x') - \zeta_i(x)|>\tau} \nonumber \\
    & ~~~+ \lambda \cdot \max_{j \in \mathcal{Y}\setminus y} \{~[f(x')]_j - [f(x)]_y~\} \nonumber \\
    & ~~~+ \eta \cdot p_{\text{LM}}(x') ~-~\nu \cdot d_{Lev}(x',x),
\end{align}
where $\zeta(x)$ is a vector such that $[\zeta(x)]_i=\zeta_i(x)$, $\beta, \lambda, \eta, \nu > 0$ are hyperparameters that trade-off different aspects, and $||\cdot||_{\infty}$ is the $l_{\infty}$ norm. The first term in the objective function encourages to pick an $x'$ where at least one attribute is either added/removed from $x$. The second term minimizes the number of such attributes for ease of interpretation. The third term is the contrastive score, which encourages the perturbed instance to be predicted different than the original instance. Fourth and fifth terms ensure that the contrast is fluent and close to the original instance, respectively.

The above objective function defines a controlled natural language generation problem. Earlier methods for controlled generation that shift the latent representation of a language model (such as GPT-2) \cite{gyc, pplm} have resulted in generated sentences being very different from the original sentence. We thus adopt a different strategy where we first take the original sentence and identify locations where substitutions/insertions need to be made using available feature attribution methods such as Integrated Gradients \cite{integrated_gradients}. 
These words are ordered by their attribution and greedily replaced with a \texttt{[MASK]} token. An MLM pre-trained BERT model \cite{transformer, bert} is then used to fill these masks. We take the top $k$ such replacements ranked by BERT likelihood. For insertions, a mask token is inserted to the right and left of important words in order to generate a set of perturbations similar to the input example. The attribute classifiers are applied to each generated candidate contrast, and the best $x'$ is selected as evaluated by Eq.~\ref{eqn:main}. For $m$ token perturbations, the above process is repeated $m$ times, where at each round, the top $k$ perturbed texts are ranked and selected according to Eq.~\ref{eqn:main}, and the above perturbation process is applied to all selected perturbed texts from the previous round. Note that we perform the hyperparameter tuning for Eq.~\ref{eqn:main} only once per dataset. Details on hyperparameter tuning and optimizing Eq.~\ref{eqn:main} are in Appendix \ref{sec:hyperparam}.

Regarding generalizability of our approach, as already noted, the attribute classifiers can be derived from other sources of data and are not necessarily dependent on the data and model being explained. Furthermore, other methods could be used to obtain attributes; unsupervised methods such as LDA, VAEs, GANs could be leveraged to ascertain semantically meaningful attributes. The attribute classifiers that appear in the loss function of Eq.~\ref{eqn:main} could be replaced by disentangled representations learned by VAEs \cite{DIP-VAE} or by topic models. Hence, CAT is generalizable beyond annotated datasets.

%
\section{Experimental Study}
\label{sec:exp}


\subsection{Setup Details}
We use an MLM pre-trained BERT\footnote{\url{https://huggingface.co/bert-base-uncased}} model from Huggingface \cite{huggingface} to generate text perturbations. For attributes, classes from the Huffpost News-Category \cite{news_category} and 20 Newsgroups \cite{kaggle20NewsGrp} datasets were used. The Huffpost dataset has 200K news headlines split into 41 classes. We merged similar classes and removed those which weren't a standard topic; 22 classes remained. The 20 Newsgroups dataset has 18000 newsgroup posts with 20 topic classes. Together, we obtained 42 attributes. For 22 classes from Huffpost, we trained 22 1-vs-all binary classifiers with a distilbert \cite{distilbert} base, so that the same sentence can have multiple classes. For 20 Newsgroups, we trained multiclass classifiers on the other 20 classes. More details on attribute classifiers are provided in Appendix \ref{sec:attr}. Note that attribute classifiers are transferable as they need not depend on the dataset and model being explained.

We evaluate our explanation method on models trained on AgNews \cite{AgNews2015}, DBPedia \cite{Lehmann2015DBpediaA}, Yelp \cite{Shen2017}, and NLI \cite{nli}\eat{, and Hate Speech \cite{hatespeech}}. For an apples-to-apples comparison of our methods with GYC on AgNews, DBPedia and Yelp, we trained models with the same architecture as the ones in their work: an Embedding Bag layer followed by a linear layer. 
\eat{Ideally we would like to compare our model with contemporary work POLYJUICE \cite{polyjuice} and MICE \cite{mice} as well.  
POLYJUICE generates counterfactual based on human in the loop control codes and not based on classifier's behaviour which we aim to explain. 
Thus our method is very different from what POLYJUICE is designed for.} 
%
For MICE the Roberta based model was used for all datasets as that is what the publicly provided implementation naturally applies to. MICE uses a two-step framework to generate conterfactual explanations, with the generator being T5 \cite{rafel-T5} fine tuned on the task-specific dataset. More details on model training are provided in Appendix \ref{sec:classifier_details} and on datasets in Appendix \ref{sec:datasets}.


\begin{table*}[t]
\centering
\small
\caption{Five examples of CAT applied to the AgNews and NLI datasets. Modifications to/in original/contrasts are shown by strikeout/red highlighting. Attribute changes offer insight into how the black box model views certain words with multiple possible meanings. For NLI, the $<$/s$>$ delimiter separates text from hypothesis.}
\label{tab:agnews}
\begin{tabular}{|p{8.25cm}|p{2.5cm}|p{2.5cm}|p{2.5cm}|}
\multicolumn{4}{c}{\textbf{AgNews}}\\
\hline
\multicolumn{1}{|c|}{\multirow{2}{*}{\textbf{Input/CAT}}} & \multicolumn{1}{c|}{\multirow{2}{*}{\textbf{Attribute Changes}}}  & \multicolumn{1}{c|}{\textbf{Input}} & \multicolumn{1}{c|}{\textbf{Contrast}} \\
   &  & \multicolumn{1}{c|}{\textbf{Pred}}  &\multicolumn{1}{c|}{\textbf{Pred}} \\ \hline

\multirow{2}{*}{Kazaa Owner Cheers \st{File} \textcolor{red}{Salary}-Swapping Decision (AP)} \quad\quad\quad\quad\quad\quad\quad\quad & \multirow{2}{*}{\shortstack{+4sale, +bball, +hockey,\\ -arts, -wndws, -cryptgphy}} & \multicolumn{1}{c|}{\multirow{2}{*}{sci-tech}} & \multicolumn{1}{c|}{\multirow{2}{*}{sports}}  \\
\hline

New Human Species \st{Discovered} \textcolor{red}{influenza} & \multicolumn{1}{c|}{+medicine, -cryptgphy} & \multicolumn{1}{c|}{sci-tech} & \multicolumn{1}{c|}{world} \\ \hline

US shows flexibility on \st{Israeli} \textcolor{red}{virtual} settlements &\multicolumn{1}{c|}{+arts, +cryptgphy, -mideast} & \multicolumn{1}{c|}{world} & \multicolumn{1}{c|}{sci-tech} \\ \hline

\eat{Will sinking Lowe resurface in \st{playoffs} \textcolor{red}{2020}? & \multicolumn{1}{c|}{+space, -hockey} & \multicolumn{1}{c|}{sports} & \multicolumn{1}{c|}{business} \\ \hline}

Pace of U.S. \st{Factory} \textcolor{red}{population} Growth Climbs in Dec & \multicolumn{1}{c|}{+politics, -money, -travel} & \multicolumn{1}{c|}{business} & \multicolumn{1}{c|}{sci-tech} \\ \hline

It may take 146 years for Nigeria to wipe out \st{corruption} \textcolor{red}{funds} from \st{its} \textcolor{red}{bank} system going by the latest report ...\eat{by Transparency International (TI) which gave the country the third position among the most corrupt countries of the world} & \multicolumn{1}{c|}{\multirow{2}{*}{+ money, - politics}} & \multicolumn{1}{c|}{\multirow{2}{*}{world}} & \multicolumn{1}{c|}{\multirow{2}{*}{business}} \\ \hline
\eat{
\multicolumn{4}{c}{}\\
\multicolumn{4}{c}{\textbf{Hate Speech}} \\
\hline
\multicolumn{1}{|c|}{\multirow{2}{*}{\textbf{Input/CAT}}} & \multicolumn{1}{c|}{\multirow{2}{*}{\textbf{Attribute Changes}}}  & \multicolumn{1}{c|}{\textbf{Input}} & \multicolumn{1}{c|}{\textbf{Contrast}} \\
   &  & \multicolumn{1}{c|}{\textbf{Pred}}  &\multicolumn{1}{c|}{\textbf{Pred}} \\ \hline

Ever try to make a \st{girl} \textcolor{red}{patient} play video games? they are terrible. these are the skills they will need for the future. we're \$\$.@ & \multicolumn{1}{c|}{\multirow{2}{*}{\shortstack{+medicine, -arts,\\  -parenting}}} & \multicolumn{1}{c|}{hate} & \multicolumn{1}{c|}{not hate}\\ \hline

It sounds so ugly to hear a \st{female} \textcolor{red}{parent} talk about another female claiming she’s a “\$\$ or \$\$” like \$\$ it ain’t yo \$\$ why u worried for& \multicolumn{1}{c|}{\multirow{2}{*}{\shortstack{+parenting, +atheism, \\-entertainment, -politics}}} &\multicolumn{1}{c|}{\multirow{2}{*}{hate}}  & \multicolumn{1}{c|}{\multirow{2}{*}{not hate}} \\ \hline

Illegal \st{immigration} \textcolor{red}{software} drives up real estate prices, widens wealth-gaps, reduces high-tech investment, increases state and local tax burdens, hurts kids' schools and college education, pushes Americans away from high-tech jobs & \multicolumn{1}{c|}{\multirow{4}{*}{+money, +science}} &\multicolumn{1}{c|}{\multirow{4}{*}{hate}}  & \multicolumn{1}{c|}{\multirow{4}{*}{not hate}} \\ \hline
When \st{you} \textcolor{red}{women}'re out drinking all night vs when the bill comes & \multicolumn{1}{c|}{+politics, -environ., -guns} &\multicolumn{1}{c|}{not hate}  & \multicolumn{1}{c|}{hate}  \\
 \hline

\eat{I thought you \textcolor{red}{ladies} were leaving the country you washed up \$\$\$\$ \$\$\$\$&+politics &\multicolumn{1}{c|}{not hate}  & \multicolumn{1}{c|}{hate}  \\ \hline}

\multirow{2}{*}{Readers respond: Kudos on \textcolor{red}{illegal} immigrant \st{features} \textcolor{red}{women} \quad\quad\quad\quad\quad\quad} &\multicolumn{1}{c|}{\multirow{2}{*}{\shortstack{+latino voices, +politics\\-travel, -mideast}}} &\multicolumn{1}{c|}{\multirow{2}{*}{not hate}} & \multicolumn{1}{c|}{\multirow{2}{*}{hate}}  \\ 
\multicolumn{1}{|c|}{}&\multicolumn{1}{c|}{}&\multicolumn{1}{c|}{}&\multicolumn{1}{c|}{}\\ \hline 
}
\multicolumn{4}{c}{}\\
\multicolumn{4}{c}{\textbf{NLI}} \\
\hline
\multicolumn{1}{|c|}{\multirow{2}{*}{\textbf{Input/CAT}}} & \multicolumn{1}{c|}{\multirow{2}{*}{\textbf{Attribute Changes}}}  & \multicolumn{1}{c|}{\textbf{Input}} & \multicolumn{1}{c|}{\textbf{Contrast}} \\
   &  & \multicolumn{1}{c|}{\textbf{Pred}}  &\multicolumn{1}{c|}{\textbf{Pred}} \\ \hline

Two outdoor workers conversing while on \st{break} \textcolor{red}{lawn}. $<$/s$>$ people on sidewalk &\multicolumn{1}{c|}{\multirow{2}{*}{-electronics}} &\multicolumn{1}{c|}{\multirow{2}{*}{\shortstack{entail\\-ment}}} & \multicolumn{1}{c|}{\multirow{2}{*}{\shortstack{contra\\-diction}}} \\ \hline

The double sink is freshly polished chrome. $<$/s$>$ The sink is \st{chrome} \textcolor{red}{rust} colored&\multicolumn{1}{c|}{\multirow{2}{*}{\shortstack{+space, -graphics \\-electronics}}}
 &\multicolumn{1}{c|}{\multirow{2}{*}{\shortstack{entail\\-ment}}} & \multicolumn{1}{c|}{\multirow{2}{*}{\shortstack{contra\\-diction}}}\\ \hline

A group of children, wearing white karate shirts, look at the American flag.$<$/s$>$ The children \st{are} \textcolor{red}{looked} at a karate tournament.&\multicolumn{1}{c|}{\multirow{2}{*}{-entertainment}} &\multicolumn{1}{c|}{\multirow{2}{*}{neutral}} & \multicolumn{1}{c|}{\multirow{2}{*}{\shortstack{contra \\ -diction}}} \\ \hline

A man is about to \st{play} \textcolor{red}{steal} his guitar.$<$/s$>$ a man is performing for school children& \multicolumn{1}{c|}{\multirow{2}{*}{-entertainment}}&\multicolumn{1}{c|}{\multirow{2}{*}{neutral}} & \multicolumn{1}{c|}{\multirow{2}{*}{\shortstack{contra \\ -diction}}} \\ \hline

Two women are giving each other a hug while a man holding a glass is looking at the camera. $<$/s$>$ The people are all taking \st{naps}  \textcolor{red}{photos} during the hottest part of the day. & \multicolumn{1}{c|}{\multirow{3}{*}{+graphics, -electronics}}& \multicolumn{1}{c|}{\multirow{3}{*}{\shortstack{contra \\ -diction}}} & \multicolumn{1}{c|}{\multirow{3}{*}{neutral}} 
\\ \hline

\end{tabular}
 \end{table*}

\subsection{Qualitative Evaluations}

We now provide qualitative examples from two datasets, AgNews and NLI, with additional examples \eat{from these,} for Yelp, and DBpedia in the Appendix \ref{sec:qualitative_examples}.

\noindent\textbf {AgNews}. The dataset is from a real-world news domain which contains short news headlines and bodies from four news categories - world, business, sports, and sci-tech. 
Our experiments focus on explanations for predicting the class from headlines.

Table \ref{tab:agnews} (top) shows results of applying CAT to five headlines in the AgNews dataset. 
The first row explains that the headline is predicted as sci-tech because if the headline was more related to topics such as things for sale, baseball, hockey, and less about computer-related topics, it would have been predicted sports, which is achieved in the contrast by replacing ``File" with ``Salary". It is important to consider the interaction of words; here, the black box model considers Kazaa a sports team because of the change. The second and third rows offer intuitive examples as to how the attribute changes relate to the contrasts. The insight is that we learn what topics the black box model finds most relevant to the prediction, as opposed to only knowing the single word and needing to figure out why that caused the change. \eat{The fourth row is not as intuitive, but digging deeper, we learned that replacing ``playoffs" by ``2019", ``2020", or ``2021" gave the same attribute changes and hence years are associated with the topic space, likely due to an interaction between the word ``resurface" and the years. Adding such a topic can still lead to an orthogonal contrast class such as business as the black box model associates ``Lowe" with the home improvement store after removing the word ``playoffs".} In the fourth row, adding politics and removing money leads to changing the input from business to sci-tech as ``factory growth" has more relationship to money while ``population growth" is related to politics. The fifth row shows the opposite as adding money and removing politics changes the input from world to business. This last example illustrates that, for longer text, single perturbations are often insufficient to flip the label and multiple changes are needed. These last two examples offer the insight that the classifier associates business with politics, which is not obvious a priori.
\eat{
\noindent\textbf{Hate Speech}. The dataset is composed of hate speech tweets labelled as hate or non-hate. It contains about 8100 tweets for training and 900 tweets for testing. The text in such tweets is much less structured and of much poorer grammar than the other datasets we experiment with. Nevertheless, CAT offers interesting insight as seen in Table \ref{tab:agnews} (middle). Note that we have hidden vulgar words (only for presentation purposes). 

The first two rows illustrate the black box model's potential bias of relating hate speech with female gender tweets; when the tweets are modified from pertaining to females, the tweets are predicted as not hate. Different topics are used to modify the tweets such as medicine where the contrast discusses a patient of unknown gender. The third row illustrates insight into the model that requires an interaction of words. The word ``immigration" by itself is not necessarily aligned with hate speech but ``illegal immigration" is more likely aligned with such speech as shown by the prediction. Adding attributes about money and science are used to explain how the tweet can be changed to not hate. The last two rows make the same point as the first two rows but in the reverse direction; modifying tweets that are not hateful by adding female connotations can easily change the prediction to hate (as gender discrimination is a form of hate speech). This is a potential bias of the black box model that should be investigated and is how one can benefit from using contrastive explanations. The insight from the attributes is that there are different topics through which the bias can be introduced. Moreover, the last example shows again that a single perturbation is not enough; in this case, an insertion and replacement are required to find a contrast showing that while the addition of ``women" biases the classifier to predict hate, the interaction of words is important as an additional insertion is required.
}

\noindent\textbf{NLI}. The Natural Language Inference \cite{nli} dataset contains samples of two short ordered texts, and the labels are either contradiction if the second text contradicts the first, neutral if the two texts do not contradict or imply one another, or entailment if the second text is a logical consequence of the first text.

Table \ref{tab:agnews} (bottom) illustrates CAT applied to five example texts from the NLI dataset. The first row shows an entailment that was modified to a contradiction by replacing the word ``break" with ``lawn". While this would be the explanation offered by a typical counterfactual method, CAT additionally shows that the topic of electronics (often associated with the word ``break") was removed. Such insight can offer more clarity as to why the contrast was predicted a contradiction rather than neutral, which seems more likely until we learn that the electronics topic that has nothing to do with the hypothesis was removed from the text. In row two, the change of ``chrome" to ``rust" is attributed to adding the space topic (as rust is an important issue in space exploration) and removed the graphics and electronics topics associated with chrome (books or web browsers). In row three, the difference of children being a part of a tournament versus simply watching the tournament is attributed to a reduction of the entertainment topic (and similarly in row four for playing a guitar versus stealing a guitar). In row five, the change of ``naps" to ``photos" is attributed to adding the graphics topic, which makes sense and helps build trust in the model.

\eat{
\textbf{DBPedia}. This dataset is derived from Wikipedia which focuses mainly on location and organizations of 14 classes. The dataset contains 560K training and 70K test examples. 

\textbf{Yelp}. The dataset is composed on informal text containing reviews of business of two classes, namely positive and negative. We use a filtered version of YELP Polarity dataset by choosing sentences of length upto 15 words. The filtered dataset contains 250K negative and 350K positive ones. 
}

\begin{table}[t]
 \centering
\caption{We evaluate CAT on four properties: 1) \textbf{Flip} rate, ii) \textbf{Dist}(distance), iii) \textbf{Fluency}, iv) \textbf{Cont} (content preservation).  We report mean values for each metric. $\uparrow$ ($\downarrow$) indicates the higher (lower) desired score wheareas $\approx 1$ indicates desired value be close to $1$. Best results are bolded and NR implies no result. Differences are statistically significant; see Appendix \ref{sec:quant_stat}.
}
 \label{tab:quant}
  \resizebox{7.5cm}{!} {\begin{tabular}{|c|c|c|cccc|}
	\multicolumn{7}{c}{\textbf{Embedding Bag based Classifier}}\\ \hline
    \multirow{2}{*}{Dataset} & \multirow{2}{*}{Method}
       & Fine & $\uparrow$ & $\downarrow$ & $\uparrow$ & $\approx 1$ \\ 
       & & Tuning & Flip & Dist & Cont & Fluency \\ \hline

	\multirow{2}{*}{AgNews} & GYC & No & 0.42 & 0.40  & 0.77 & 1.13 \\
	& CAT & No & \textbf{1} & \textbf{0.22}  & \textbf{0.87} & \textbf{1.01} \\ \hline
	
	\multirow{2}{*}{Yelp}& GYC & No & 0.70 & 0.49 & 0.61 & 1.32 \\ 
	& CAT &No & \textbf{1} & \textbf{0.23} & \textbf{0.88} & \textbf{1.09} \\ \hline
	
	\multirow{2}{*}{Dbpedia}& GYC&No & 0.72&  0.52 &0.55 & 1.33 \\
	& CAT &No & \textbf{1} & \textbf{0.16} &\textbf{ 0.89} & \textbf{1.05} \\ \hline
	
	\multirow{2}{*}{NLI}& GYC &No & \multicolumn{4}{c|}{NR}\\
	& CAT &No & \textbf{1} &\textbf{0.08} &\textbf{0.98} & \textbf{1.03} \\ \hline

	\multicolumn{7}{c}{}\\
	\multicolumn{7}{c}{\textbf{Roberta based Classifier}}\\ \hline

    \multirow{2}{*}{Dataset} & \multirow{2}{*}{Method}
       & Fine & $\uparrow$ & $\downarrow$ & $\uparrow$ & $\approx 1$ \\ 
       & & Tuning & Flip & Dist & Cont & Fluency \\ \hline

	\multirow{3}{*}{AgNews} & MICE & Yes & 1 & 0.42  & 0.79 & \textbf{1.01} \\
	 & MICE-nft & No & 0.98 & 0.52  & 0.70 & 0.90 \\
	& CAT & No & \textbf{1} & \textbf{0.19}  & \textbf{0.90} & 0.90 \\  
	 \hline
\multirow{3}{*}{Yelp} & MICE & Yes & 0.99 & 0.35  & 0.86 & 1.09 \\
	 & MICE-nft & No & 0.97 & 0.41  & 0.82 & 1.06 \\
	& CAT & No & \textbf{1} & \textbf{0.11}  & \textbf{0.96} & \textbf{1.04} \\ 
	 \hline
	 \multirow{3}{*}{Dbpedia} & MICE & Yes & 0.81 & 0.28  & 0.90 & 1.23 \\
	 & MICE-nft & No & 0.75 & 0.34  & 0.84 & 1.19 \\
	& CAT & No & \textbf{1} & \textbf{0.09}  & \textbf{0.96} & \textbf{1.06} \\ 
	 \hline
	 	 \multirow{3}{*}{NLI} & MICE & Yes & 1 & 0.25  & 0.85 & 1.14 \\
	 & MICE-nft & No & 0.99 & 0.29  & 0.90 & 1.12 \\
	& CAT & No & \textbf{1} & \textbf{0.07}  & \textbf{0.98} & \textbf{1.03} \\ 
	 \hline
\end{tabular}}
  \vspace{-0.2cm}
\end{table}

\eat{\begin{table*}[htbp]
 \centering
 \small
\caption{We evaluate CAT on four properties: 1) \textbf{Flip} rate, ii) \textbf{Dist}(distance), iii) \textbf{Fluency}, iv) \textbf{Cont} (content preservation).  We report mean values for each metric. $\uparrow$ ($\downarrow$) indicates the higher (lower) desired score wheareas $\approx 1$ indicates desired value be close to $1$. Best results are bolded and NR implies no result. Differences are statistically significant; see Appendix \ref{sec:quant_stat}.
}
 \label{tab:quant}
 \begin{tabular}{l|cccc|cccc|cccc}
    \toprule
    \multirow{3}{*}{Method} &
      \multicolumn{4}{c|}{AgNews } &
      \multicolumn{4}{c|}{Yelp } &
       \multicolumn{4}{c}{Dbpedia } \\
       & $\uparrow$ & $\downarrow$ & $\uparrow$ & $\approx 1$ & $\uparrow$ & $\downarrow$ & $\uparrow$ & $\approx 1$ & $\uparrow$ & $\downarrow$ & $\uparrow$ & $\approx 1$ \\
       & Flip & Dist & Cont & Fluency &
       Flip & Dist & Cont & Fluency &
       Flip & Dist & Cont & Fluency\\ 
      \midrule
	GYC & 0.42 & 0.40  & 0.77 & 1.13 &0.70 & 0.49 & 0.61 & 1.32 & 0.72&  0.52 &0.55 & 1.33 \\
	CAT & \textbf{1} & \textbf{0.22}  & \textbf{0.87} & \textbf{1.01} & \textbf{1} & \textbf{0.23} & \textbf{0.88} & \textbf{1.09} & \textbf{1} & \textbf{0.16} &\textbf{ 0.89} & \textbf{1.05}\\
    \bottomrule
\end{tabular}
\begin{tabular}{l|cccc|cccc}
\multicolumn{9}{c}{}\\
\multirow{3}{*}{Method} &
     \multicolumn{4}{c|}{NLI} &
      \multicolumn{4}{c}{Hate Speech} \\
      & $\uparrow$ & $\downarrow$ & $\uparrow$ & $\approx 1$ & $\uparrow$ & $\downarrow$ & $\uparrow$ & $\approx 1$ \\
      & Flip & Dist & Cont & Fluency &
       Flip & Dist & Cont & Fluency\\
        \midrule
       GYC & \multicolumn{4}{c|}{NR} & \multicolumn{4}{c}{NR}\\
       CAT &\textbf{1} &\textbf{0.08} &\textbf{0.98} & \textbf{1.03}&\textbf{1} &\textbf{0.07} &\textbf{0.97} &\textbf{1.02} \\
      \bottomrule
  \end{tabular}
  \label{tab:quant}
  \vspace{-0.2cm}
\end{table*}
}
\subsection{Quantitative Evaluations}
We evaluate the explanation methods on 500 randomly selected test instances from each dataset. We do not report results on NLI for GYC as it did not produce valid contrasts possibly because of the longer length of the texts\eat{ and incompatibility with the Roberta classifier}. For each dataset, we measure the following properties: i) \textit{Flip rate} (Flip), ii) \textit{Edit distance} (Dist), iii) \textit{Content Preservation} (Cont), and iv) \textit{Fluency}. Flip rate is a measure of the model's efficacy to generate contrastive sentences and is \eat{i.e. the generated sentences results in change of predicted class due to word edits.} defined as the fraction of inputs for which an edit successfully flips the prediction. Edit Distance is the number of edits as measured by the word-level Levenstein distance between input and contrastive sentences, i.e., the minimum number of deletions, insertions, or substitutions required to transform one into the other. We report a normalized version given by the Levenshtein distance divided by the number of words in the input; this metric ranges from 0 to 1. Content preservation measures how much input content is preserved while generating a contrastive sentence in a latent embedding space. For this, we compute the cosine similarity between input and contrastive sentence embeddings obtained from a pre-trained sentence BERT \cite{sentence-BERT} model. Fluency measures the alignment of the contrastive and input sentence distributions. We evaluate fluency by calculating masked language modeling loss on both original and edited sentences using a pre-trained GPT-2 model and compute fluency as the ratio of the loss of the contrast to the loss of the input sentence.\eat{ The average loss obtained across sentence pairs are reported as fluency.} A value of 1.0 indicates the contrast is as fluent as the original sentence.
Table \ref{tab:quant} reports means of each metric across all instances obtained from generated contrastive sentences.  

\begin{figure*}[htbp]
\hspace{-.3cm}
\begin{tabular}{l}
   \includegraphics[width=.24\textwidth]{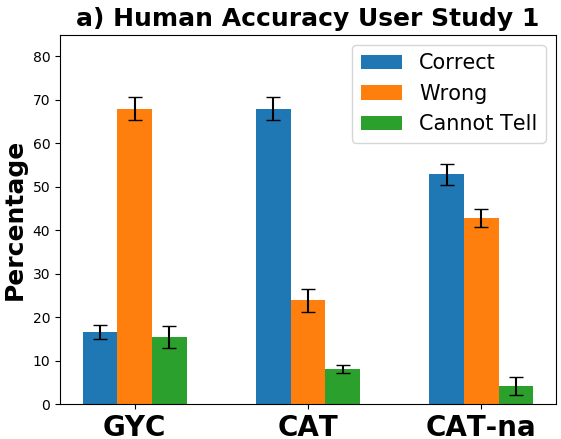}   
   
   \includegraphics[width=.24\textwidth]{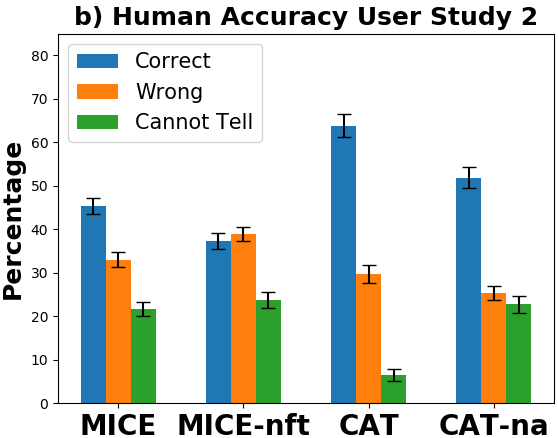}   
   
   \includegraphics[width=.24\textwidth]{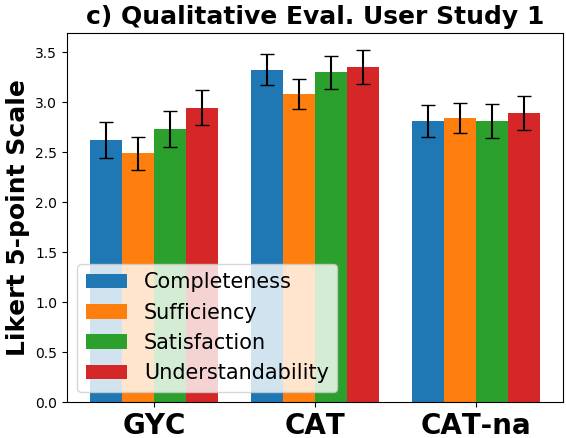}
   
   \includegraphics[width=.24\textwidth]{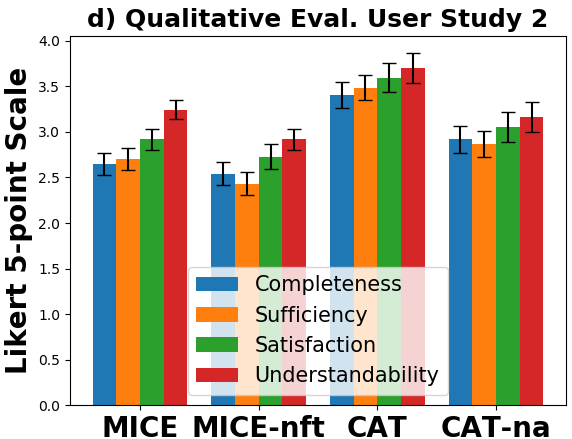}
\end{tabular}
   \vspace{-.25cm}
  \caption{Figures a) and b) show the percentage (human) accuracy in predicting the label of the input sentence based on the contrastive explanations. Users perform significantly better using our CAT explanations showcasing the benefit of the attributes. Figures c) and d) show a 5-point Likert scale (higher better) for four qualitative metrics used in previous studies \cite{actinfl} compared across the methods. Here too, the difference is noticeable for all four metrics. Error bars are one standard error.}
  \label{fig:userstudy}
\end{figure*}
\noindent\textbf{Observations:} 
In Table \ref{tab:quant}  we compare the performance of CAT with two state-of-the-art contrastive methods GYC and MICE. MICE-nft denotes MICE without fine tuning. As can be seen CAT produces contrasts with perfect flip rate, retains highest content relative to the original sentence, that too with fewest changes, and maintains best language fluency in all cases, but one. 
This can be accredited to the core details of the CAT approach which is based on minimal but relevant perturbations through a controlled local greedy search procedure guided by attribute classifiers. The single case where CAT is not best performing is possibly because fine tuning helps create more natural contrasts; also confirmed by the similar performance of MICE-nft and CAT. 

We also estimated the efficiency of CAT by computing the time it takes to obtain a contrastive explanation normalized by the input length. CAT, MICE, and GYC were evaluated on a NVIDIA V100 GPU for 10 contrastive explanations. The mean times taken by CAT, MICE and GYC were $2.37$, $1.17$ and $10.69$ seconds respectively. The efficiency of CAT over GYC can be credited to the controlled local greedy search approach, although guiding the search with attribute classifiers makes it slightly more expensive than MICE, where for the latter we are ignoring the time for fine tuning.

\subsection{Human Evaluation}
\label{sec:human}
We now describe two user studies we conducted to ascertain the value of our attributed explanations. User studies have become ubiquitous in explainable AI literature \cite{lime, acd, mame, CEM-MAF} because they illustrate the benefit of these tools to the end user. We follow in the direction of most user studies by ascertaining benefit by requiring participants to perform a task  given our explanations\eat{, as opposed to solely asking users the utility of our method}.

\noindent\textbf{Methods:} We consider five different explanation methods: 1) CAT, 2) CAT-na (i.e. CAT without attributes), 3) GYC, 4) MICE and 5) MICE-nft (i.e. MICE without fine-tuning). Comparison of CAT with CAT-na provides an ablation, while its comparison with GYC, MICE, MICE-nft showcases its value relative to other state-of-the-art contrastive text explanations. Again, here we ran two separate user studies, one comparing with GYC (User Study 1) and the other with MICE (User Study 2), since each implementation was more suited to a specific type of model and embedding. CAT, however, was amenable to either setting.

\noindent\textbf{Setup:} We chose the AgNews dataset for the study since this was the only dataset where we under-performed on one of the benchmark metrics (fluency) w.r.t. the competitors. We thus wanted to see if this had any effect in terms of the quality of insight provided by our explanations to typical users of such explanations. For each study, we built a four class neural network black box model (described earlier) to predict articles as either belonging to Business, Sci-Tech, World or Sports categories. The task given to users was to determine the classification of the article based on a contrastive explanation from one of the respective methods. Five options were provided to the user which included the four classes along with a ``Can't Tell" option. For each method, seven (User Study 1) or five (User Study 2) randomly chosen sentence-explanation pairs were provided to the user where the users were blinded to the exact method producing the explanation. For each (anonymized) explanation method, we additionally asked the users for their qualitative assessment along four dimensions; completeness, sufficiency, satisfaction and understandability based on a 5 point Likert scale. A total of 24 questions were answered by users in both user studies. Users were also allowed to leave optional comments which we provide in the appendix, along with screen shots from the user study. 

We hosted our survey on Google Forms. A total of 75 participants with backgrounds in data science, engineering and business analytics voluntarily took part in user studies (37 in study 1 and 38 in study 2). We 
chose this demographic as recent studies show that typical users of such explanations have these backgrounds \cite{umang}.

\noindent\textbf{Observations:} 
Figure \ref{fig:userstudy} depicts the results from our user studies. Figure \ref{fig:userstudy}a demonstrates that both our methods CAT and its ablation CAT-na (which does not use attributes) significantly outperform GYC in terms of usability towards the task.
Figure \ref{fig:userstudy}b shows similar performance for CAT and CAT-na, although MICE is a much stronger competitor than GYC. It seems that the embeddings used by MICE lead to better contrasts than GYC, not to mention the mechanistic difference between them where GYC biases towards changing the end of sentences which may not be preferable in many cases. Additionally, the fine tuning done in MICE seems to further help elevate the quality of the contrasts. Nonetheless, CAT still outperforms MICE without the (expensive) need to fine tune.  

Our method performs much better than GYC, MICE and CAT-na from a qualitative perspective, as well, as seen in Figures \ref{fig:userstudy}c and \ref{fig:userstudy}d, where CAT seems to score highest in understandability. These qualitative preferences of CAT are further confirmed through the (optional) comments written by some participants, e.g., ``... explainer B was very good, and explainer C was reasonably good'', where ``explainer B'' refers to CAT and ``explainer C'' to CAT-na in user study 1; or ``the additional info in explanation C was useful'' in user study 2, where ``explanation C'' here refers to CAT. 

\section{Conclusion}
\label{s:conclusion}

In this paper, we proposed a contrastive explanation method, CAT, with a novel twist where we construct attribute classifiers from relevant subtopics available in the same or different dataset used to create the black box model, and leverage them to produce high quality contrasts. The attributes themselves provide additional semantically meaningful information that can help gain further insight into the behavior of the black box model. We have provided evidence for this through diverse qualitative examples on multiple datasets, a carefully conducted user study and through showcasing superior performance on four benchmark quantitative metrics. 

In the future, it would be interesting to test CAT on other applications with the appropriate attribute classifiers. Another useful direction is generating counterfactually-augmented data (CAD) using CAT. If used to generate multiple contrasts, CAT could offer attributional/topic diversity. Since diversity in types is key to producing robust models, CAT could potentially to a large degree alleviate ``the lack of perturbation diversity that limits CAD's effectiveness" \cite{cad}.

\section*{Limitations}
Although our work has the potential to have a positive impact on discovering models that have unknown discrimination, a nefarious agent could potentially provide accurate but purposely incorrectly labeled attribute classifiers in order to drive certain misleading or simply incorrect insights. Another unethical act by a developer could be to hide sensitive attributes so that biases could not be discovered. In order to use the explanation system in a beneficial manner we do assume the developer can be trusted. It is also possible that we may not uncover the globally minimal contrast since the optimization is non-convex given the complexity of classifiers we are trying to explain. Thus, smaller edits may be possible that change the output of a classifier that go unnoticed. This however, is a concern for even other contrastive/counterfactual explainability methods.

\bibliography{cat}
\bibliographystyle{acl_natbib}

\appendix
\section*{Appendix}

\section{Hyperparemeter tuning for CAT}\label{sec:hyperparam}
Since no ground truth exists for our contrastive explanations, hyperparameters for CAT objective (given in Eq.\ref{eqn:main}) were tuned qualitatively by observation. It is important to consider that, in practice a user will only need to tune the hyperparameters once at the beginning and will then be able to generate explanations for an arbitrary number of instances from the dataset. Our tuning led to the following values which we used across datasets: We kept highest weights for $\lambda = 5.0$ to make sure that perturbed sentences have a label different from that of original sentence. We used mean of BERT logits of words inserted/replaced as a surrogate for $p_{LM}(x')$ with regularization parameter $\eta$ set to be $1.0$ and regularization parameter $\nu$ for Levenshtein distance set to $2.0$. $\beta$ was set to $3.0$. The threshold for predicting addition or removal of attributes $\tau$, we used two values. For binary attribute classifiers from the Huffpost News dataset, we set $\tau = 0.3$, and for multiclass attribute classifiers trained on the 20 Newsgroup dataset, we set $\tau=0.05$.

We took 50 randomly selected examples from the training set, generated explanations and evaluated them manually to choose hyperparameters which is similar to prior works \cite{CEM-MAF, gyc}. The boundary values for hyperparameter search were $\lambda \in [4,10]$, $\beta \in [1,5]$, $\eta \in [0.5, 2]$, $\nu \in [1,4]$, and $\tau \in [0.01, 0.5]$.
With more examples for tuning we would expect better explanations, although we found this number to be sufficient.

\section{Attribute classifiers}\label{sec:attr}
In our experiments we created 42 attributes from Huffpost News-Category dataset\cite{news_category} and 20 Newsgroups \cite{kaggle20NewsGrp}.

News-Category dataset \footnote{\url{https://www.kaggle.com/rmisra/news-category-dataset}} has 41 classes of which we merged similar classes and removed those which weren't standard topic. For instance, classes ``\textit{food \& drink}" and ``\textit{taste}" was merged and labelled as a new class ``\textit{food}". At the end we obtained the following 22 classes: \textit{education, money, world, home \& living, comedy, food, black voices, parenting, travel, sports, women, religion, Latino voices, weddings, entertainment, crime, queer voices, arts, politics, science, fifty}, and \textit{environment}. We created 22 1-vs-all binary classifiers, so that the same sentence can have multiple classes. To alleviate the class imbalance issue in training these binary classifiers we sampled fewer instances uniformly at random from the negative classes making sure that the number of negative instances are no more than $80\%$ of the training data.

The 20 Newsgroup dataset \footnote{\url{https://scikit-learn.org/stable/modules/generated/sklearn.datasets.fetch_20newsgroups.html}} has the following 20 classes: \textit{atheism, graphics, ms-windows.misc, computer, mac.hardware, ms-windows.x, forsale, autos, motorcycles, baseball, hockey, cryptography, electronics, medicine, space, christian, guns, mideast, politics}, and \textit{religion}. A distilbert multi-class classifier was trained on this data.

All attribute classifiers were trained with DistliBERT base \footnote{\url{https://huggingface.co/distilbert-base-uncased}} for $10$ epochs using the Adam optimizer with a learning rate of $5\times 10^{-5}$, weight decay of $0.01$, and a batch size of $16$. Each of the models were trained using NVIDIA V100 GPUs in under 12 hours.

\section{Text classifiers}\label{sec:classifier_details}
We conducted experiments on two sets of text classifiers trained on four datasets: One set to compare CAT with GYC and another to compared CAT with MICE. All experiments were run on NVIDIA V100 GPUs.

For experiments comparing CAT with GYC\cite{gyc}, we trained model with the same architecture as the one used in their paper. This model is composed of an \textit{EmbeddingBag} layer followed by a linear layer, a \textit{ReLU} and another linear layer to obtain the logit. The logit is provided to a classifier we trained with cross entropy loss. A GPT2 tokenizer \footnote{\url{https://huggingface.co/gpt2-medium}} was used to convert text into bag of words. The models were trained using the Adam optimizer with a learning rate of $1\times 10^{-3}$, weight decay of $0.01$, and a batch size of $32$.

For experiments comparing CAT with MICE\cite{mice}, we used author provided implementation\footnote{\url{https://github.com/allenai/mice}} to train models with \texttt{AllenNLP} \cite{allennlp}. This architecture is composed of a \textit{RoBERTa-base} model with a linear layer. For all four datasets the models were trained for $5$ epochs with batch size of $8$ using Adam optimizer with a learning rate of $4e-05$, weight decay of $0.1$, and slanted triangular learning rate scheduler with cut frac $0.06$.

\section{Datasets}\label{sec:datasets}
We performed experiments on 4 datasets: AgNews\cite{AgNews2015}, NLI\cite{nli}, DBpedia \cite{Lehmann2015DBpediaA}, and Yelp\cite{Shen2017}. AgNews dataset was taken from Kaggle website \footnote{\url{https://www.kaggle.com/datasets/amananandrai/ag-news-classification-dataset}} and rest three datasets from \textit{huggingface datasets}\cite{hf_datasets}.

\textbf {AgNews}. The dataset is from a real-world news domain which contains short news headlines and bodies from four new categories - world, business, sports, and sci-tech. It contains ~30K training and ~1.9K test examples per class, comprising of ~128K samples. Our experiments focus on explanations for predicting the class from the headlines.

\textbf{NLI}. The Natural Language Inference \cite{nli} dataset\footnote{\url{https://huggingface.co/datasets/snli}} contains samples of two short ordered texts, and the labels are either \textit{contradiction} if the second text contradicts the first, \textit{neutral} if the two texts do not contradict or imply one another, or \textit{entailment} if the second text is a logical consequence of the first text. The dataset contains ~550K training, 10K test and 10K validation examples.

\textbf{DBpedia}. This dataset is a subset of original DBpedia data which is a crowd-sourced community effort to extract structured information from Wikipedia. This dataset\footnote{\url{https://huggingface.co/datasets/dbpedia_14}} is constructed by picking 14 non-overlapping classes from DBpedia 2014 with 40K training and 5K test examples per class. Task here is to predict the class a DBpedia entry belong to. In our experiments we only use \textit{content} and drop the \textit{title} field provided with with the dataset.

\textbf{Yelp}. This is a binary sentiment classification dataset\footnote{\url{https://huggingface.co/datasets/yelp_polarity}} containing 560K highly polar reviews for training and 38K for testing. The dataset consists of reviews from Yelp which is extracted from the Yelp Dataset Challenge 2015.

\begin{table*}[!]
 \centering
\caption{We evaluate CAT on three metrics:  i) \textbf{Dist}(distance), ii) \textbf{Fluency}, iii) \textbf{Cont} (content preservation).  We report p-values from a pairwise t-test of the difference of means between GYC with CAT and MICE with CAT for all metrics. As can be seen by the negligible p-values, CAT is (statistically) significantly better than GYC as well than MICE. We also report standard deviation (std. dev.) for all metrics for CAT, GYC and MICE, which together with the means reported in the main paper, are used to run the t-tests.
}
 \label{tab:quant_stat}
\begin{tabular}{l|ccc|ccc}
\multicolumn{7}{c}{}\\
	\multicolumn{7}{c}{\textbf{Embedding Bag based Classifier}}\\ 
\hline
\multirow{3}{*}{Stat} &
     \multicolumn{3}{c|}{AgNews} &
      \multicolumn{3}{c}{Yelp} \\
       & Dist & Cont & Fluency &
       Dist & Cont & Fluency\\
        \midrule
      p-value   &$< 1e^{-8}$  &$< 1e^{-8}$  & $< 1e^{-8}$   &$< 1e^{-8}$  &$< 1e^{-8}$  &$< 1e^{-8}$  \\
      \midrule
      std. dev. (GYC)   &0.164   &0.110   &0.286  &0.140  &0.105  &0.467  \\
      std. dev. (CAT)   &\textbf{0.098}  &\textbf{0.084} &\textbf{0.105 } &\textbf{0.100} &\textbf{0.076} &\textbf{0.134}  \\
      \bottomrule
  \end{tabular}
  
 \begin{tabular}{l|ccc|ccc}

\multicolumn{7}{c}{}\\
\multirow{3}{*}{Stat} &
     \multicolumn{3}{c|}{Dbpedia} &
      \multicolumn{3}{c}{NLI} \\
      & Dist & Cont & Fluency &
      Dist & Cont & Fluency\\
        \midrule
      p-value   &$< 1e^{-8}$  &$< 1e^{-8}$  & $< 1e^{-8}$   &$< 1e^{-8}$  &$< 1e^{-8}$  &$< 1e^{-8}$  \\
      \midrule
      std. dev. (GYC)   &0.132   &0.115   &0.344   &\multicolumn{3}{|c}{NR}  \\
      std. dev. (CAT)    &\textbf{0.059}  &\textbf{0.069}  &\textbf{0.064} &\textbf{0.323} &\textbf{0.010} &\textbf{0.033}  \\
      \bottomrule
  \end{tabular}


\begin{tabular}{l|ccc|ccc}
\multicolumn{7}{c}{}\\
	\multicolumn{7}{c}{\textbf{Roberta based Classifier}}\\ 
\hline
\multirow{3}{*}{Stat} &
     \multicolumn{3}{c|}{AgNews} &
      \multicolumn{3}{c}{Yelp} \\
       & Dist & Cont & Fluency &
       Dist & Cont & Fluency\\
        \midrule
      p-value   &$< 1e^{-8}$  &$< 1e^{-8}$  & $< 1e^{-8}$   &$< 1e^{-8}$  &$< 1e^{-8}$  &$< 1e^{-8}$  \\
      \midrule
      std. dev. (MICE)   &\textbf{0.182}   & \textbf{0.121}   &0.196  &0.174  &0.109  &0.208  \\
      std. dev. (MICE-nft)   &0.307   &0.209   &0.224  &0.240  &0.154  &0.212  \\
      std. dev. (CAT)   &0.307  &0.209 & \textbf{0.144}  & \textbf{0.084} & \textbf{0.044} &\textbf{0.098}  \\
      
      \bottomrule
  \end{tabular}
  
 \begin{tabular}{l|ccc|ccc}
 \multicolumn{7}{c}{}\\
\multirow{3}{*}{Stat} &
     \multicolumn{3}{c|}{Dbpedia} &
      \multicolumn{3}{c}{NLI} \\
      & Dist & Cont & Fluency &
      Dist & Cont & Fluency\\
        \midrule
      p-value   &$< 1e^{-8}$  &$< 1e^{-8}$  & $< 1e^{-8}$   &$< 1e^{-8}$  &$< 1e^{-8}$  &$< 1e^{-8}$  \\
      \midrule
      std. dev.(MICE)  &0.151   &0.077   &0.168  &0.132   &0.109   &0.131  \\
      std. dev.(MICE-nft)  &0.200  &0.160  &0.160 &\textbf{0.164} &0.088 &0.143 \\
      std. dev.(CAT)   &\textbf{0.058}  & \textbf{0.039}  & \textbf{0.071} &\textbf{0.037} &\textbf{0.013} &\textbf{0.054}  \\
      \bottomrule

      
  \end{tabular}
  \label{tab:quant_stat}
\end{table*}

\section{Quantitative Evaluation Statistics}\label{sec:quant_stat}
Table \ref{tab:quant} (in the main paper) shows the performance evaluation of our proposed approach, CAT, on five different datasets and $2$ classifier models, namely Embedding Bag based classifier and Roberta based classifier models. It was noted that CAT outperformed GYC  and MICE over AgNews, Yelp and Dbpedia by statistically significant margins on respective model. Recall that GYC usage Embedding Bag based classifier model and MICE usage Roberta based classifier. Therefore, we implemented both the classifier models with CAT to compare both GYC and MICE.  We verify this statement in Table \ref{tab:quant_stat} where we report pairwise t-tests comparing the means for CAT with GYC and CAT with MICE and standard deviation of CAT, GYC, MICE for each metric and dataset.  The improvement of CAT over GYC and MICE is observed to be statistically significant across all metrics. We do not report additional statistics for the flip rate metric as CAT always produces a contrastive sentence with a flipped class label unlike GYC or MICE which sometimes fails to flip.


\section{Additional Information for User Study}
\begin{figure*}[htbp]
\centering
 \begin{subfigure}[b]{0.49\textwidth}
        \centering
        \includegraphics[width=\textwidth]{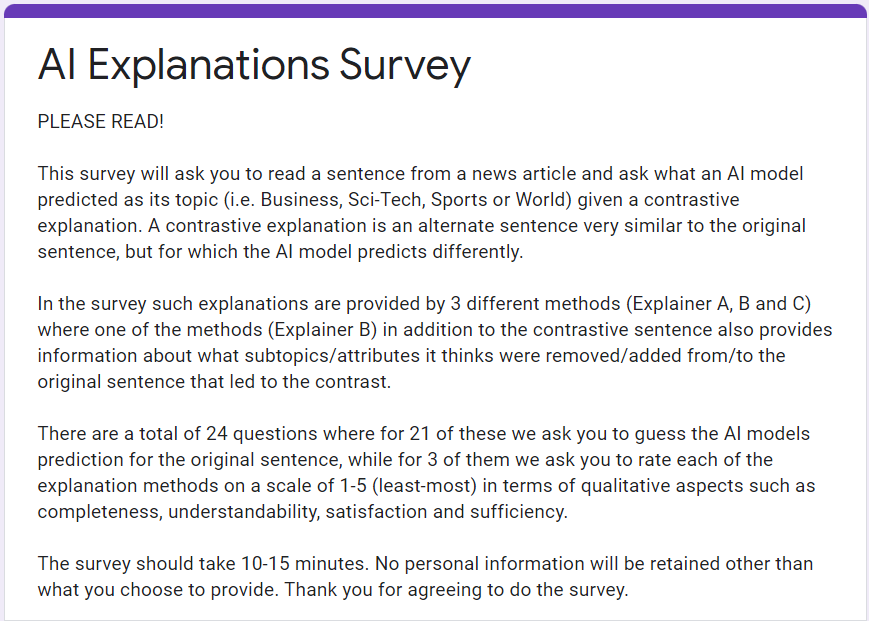}
        \caption{}
    \end{subfigure}
     \begin{subfigure}[b]{0.49\textwidth}
        \centering
        \includegraphics[width=\textwidth]{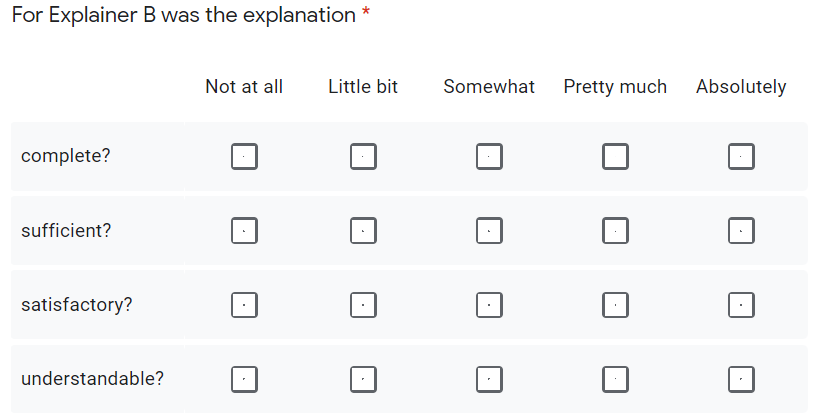}
        \caption{}
    \end{subfigure}\\
   \begin{subfigure}[b]{0.39\textwidth}
        \centering
        \includegraphics[width=\textwidth]{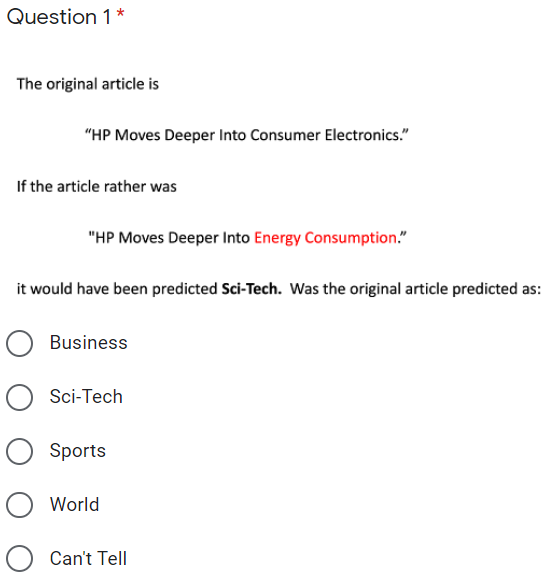}
        \caption{}
    \end{subfigure}
   \hspace{0.1\textwidth}
   \begin{subfigure}[b]{0.49\textwidth}
        \centering
        \includegraphics[width=\textwidth]{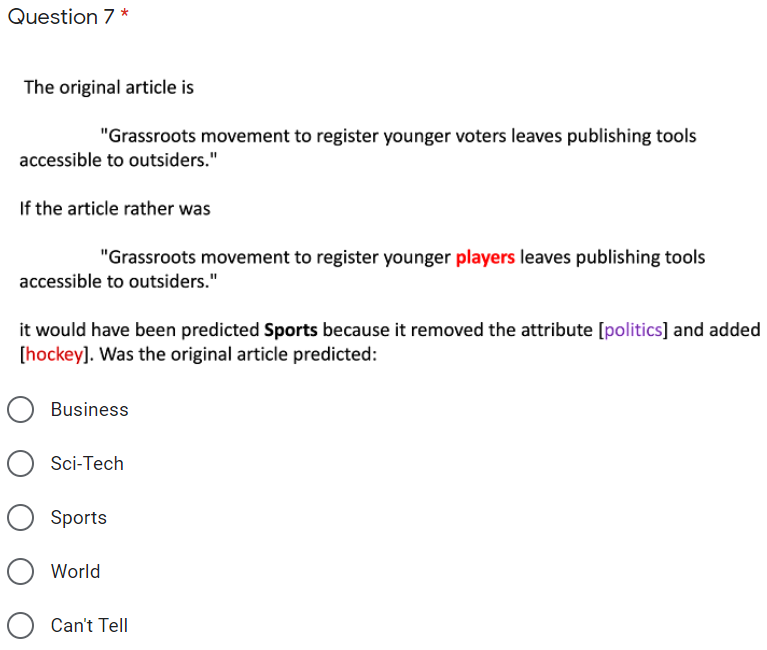}
        \caption{}
    \end{subfigure}
  \caption{In (a) we show the instructions for the survey. In (b) we see the qualitative questions. In (c) and (d) we see the task oriented questions for GYC and CAT respectively.} 
\label{fig:userinstr}
\end{figure*}

Figure \ref{fig:userinstr} shows screenshots of user study 1, including the instructions and example questions for the three different methods discussed in the study. The same instructions and format was used for user study 2, except that we asked five task oriented questions (rather than seven) per explainer as there were four explainers (as opposed to 3) keeping the total number of questions to be 24, and thus keeping the overall effort of both user studies roughly the same for participants. For the users the methods were named as Explainer A, B and C, where they correspond to GYC, CAT, and CAT-na respectively for user study 1. For user study 2, Explainer A, B, C and D corresponded to MICE, MICE-nft, CAT and CAT-na. Some users also left optional comments at the end of the user study which we list here:

\begin{itemize}
    \item ``I found the questions confusing... What is complete, sufficient and understandable explanation in a word change? Also, in each example, what was I supposed to guess, the category of the article or the possible prediction of the method? Are those things different?'' (Study 1)
    \item ``Explainer A was pretty bad, explainer B was very good, and explainer C was reasonably good.'' (Study 1)
    \item ``Explainer b was the best. The q\&a at the end of each page allows multiple choices per row, which is bad. Each question should remove the modified classification as an option (if the modified sentence is World, I shouldn't be allowed to choose World). Early on the survey should state clearly that there are only 4 categories (business/sci-tech/sports/world) which I didn't know right away, I expected each question have 4 different options, and knowing this might have helped me understand what we're doing here better. The opening text was very confusing.'' (Study 1)
    \item ``I found B to be the best one.'' (Study 1)
    \item ``nice survey'' (Study 1)
    \item ``the additional info in explanation C was useful'' (Study 2)
\end{itemize}

\section{Additional Qualitative Examples}\label{sec:qualitative_examples}
Table \ref{tab:extra_examples} offers at least five more examples of contrastive explanations from each dataset. Additional insight can be gained from the attributes; in the first row, adding travel-related text can flip the class from business to world, i.e., the article was predicted business because if it was more about travel it would have been predicted as world. This type of explanation adds extra intuition as opposed to only being given the replacement of words, ``oil" to ``winds" in this case. As can also been seen, insight is also a function of using attributes that have a relationship to the dataset and task. For example, attributes derived from news sources are often not helpful for explaining sentiment classifiers (e.g., Yelp) for which sentiment is often flipped by the negation of the text. This is to be expected; good explanations require good attributes.

\begin{table*}[htbp]
\centering
\small
\caption{Extra contrastive explanation examples from four datasets: AgNews, NLI, DBPedia, and Yelp.}
\label{tab:extra_examples}
\begin{tabular}{|p{5.75cm}|p{4cm}|p{3cm}|p{3cm}|p{1cm}|}
\hline
\multicolumn{1}{|c|}{\multirow{2}{*}{\textbf{Input/CAT}}} & \multicolumn{1}{c|}{\multirow{2}{*}{\textbf{Attribute Changes}}}  & \multicolumn{1}{c|}{\textbf{Input}} & \multicolumn{1}{c|}{\textbf{Contrast}}
& \multicolumn{1}{c|}{\multirow{2}{*}{\textbf{Dataset}}}\\
   &  & \multicolumn{1}{c|}{\textbf{Pred}}  &\multicolumn{1}{c|}{\textbf{Pred}}&  \\ \hline

\st{Oil} \textcolor{red}{Winds} Up from 7-Week Lows on U.S. Weather & \multicolumn{1}{c|}{+travel} & \multicolumn{1}{c|}{business} & \multicolumn{1}{c|}{world} & AgNews \\ \hline

New Hummer Is Smaller, Cheaper and Less \st{Gas} \textcolor{red}{cpu} Hungry & \multicolumn{1}{c|}{\multirow{2}{*}{\shortstack{+cryptography, -travel, \\-space}}} & \multicolumn{1}{c|}{\multirow{2}{*}{business}} & \multicolumn{1}{c|}{\multirow{2}{*}{sci-tech}} & \multicolumn{1}{c|}{\multirow{2}{*}{AgNews}} \\
\hline

Perfect start for France in \st{Federation} \textcolor{red}{Microsoft} Cup & \multirow{2}{*}{\shortstack{+cryptography, -motorcycles, \\-hockey, -space, -politics}} & \multicolumn{1}{c|}{\multirow{2}{*}{sports}} & \multicolumn{1}{c|}{\multirow{2}{*}{sci-tech}} & \multicolumn{1}{c|}{\multirow{2}{*}{AgNews}} \\ \hline

Will sinking Lowe resurface in \st{playoffs} \textcolor{red}{2020}? & \multicolumn{1}{c|}{+space, -hockey}& \multicolumn{1}{c|}{sports} & \multicolumn{1}{c|}{sci-tech} & AgNews \\ \hline

\st{Kuwait} \textcolor{red}{Source}: Fundamentalists Recruiting Teens (AP) &  \multicolumn{1}{c|}{+guns, religion, -mideast}& \multicolumn{1}{c|}{world} & \multicolumn{1}{c|}{sci-tech} & AgNews \\ \hline

\multirow{2}{*}{Harry \st{in nightclub} \textcolor{red}{inflight} scuffle}& \multicolumn{1}{c|}{\multirow{2}{*}{\shortstack{+sports, +space, -religion, \\-cryptography, -electronics}}} & \multicolumn{1}{c|}{\multirow{2}{*}{world}} & \multicolumn{1}{c|}{\multirow{2}{*}{sci-tech}} & \multicolumn{1}{c|}{\multirow{2}{*}{AgNews}} \\
\multicolumn{1}{|c|}{}&\multicolumn{1}{c|}{}&\multicolumn{1}{c|}{}&\multicolumn{1}{c|}{}&\multicolumn{1}{c|}{}\\ \hline
\eat{
I thought you \textcolor{red}{ladies} were leaving the country you washed up \$\$\$\$ \$\$\$\$? & \multicolumn{1}{c|}{\multirow{2}{*}{+politics, -cryptography}}& \multicolumn{1}{c|}{\multirow{2}{*}{hate}} & \multicolumn{1}{c|}{\multirow{2}{*}{not hate}} & \multicolumn{1}{c|}{\multirow{2}{*}{Hate}} \\ \hline

Such crimes must attract severest punishment,say Min Jail term of 25Y. If they r \st{illegal} \st{migrants} \textcolor{red}{convicted}, force them to leave the Country immediately& \multicolumn{1}{c|}{\multirow{3}{*}{+atheism, -politics}} & \multicolumn{1}{c|}{\multirow{3}{*}{hate}} & \multicolumn{1}{c|}{\multirow{3}{*}{not hate}} & \multicolumn{1}{c|}{\multirow{3}{*}{Hate}} \\ \hline

\multirow{2}{*}{\st{@brianstelter} \textcolor{red}{she} is a complete \st{joke} \textcolor{red}{\$\$\$\$} ! if you see...}& 
\multicolumn{1}{c|}{\multirow{2}{*}{\shortstack{+women, +cryptography, \\-sports, -politics}} }& \multicolumn{1}{c|}{\multirow{2}{*}{not hate}} & \multicolumn{1}{c|}{\multirow{2}{*}{hate}} & \multicolumn{1}{c|}{\multirow{2}{*}{Hate}} \\
\multicolumn{1}{|c|}{}&\multicolumn{1}{c|}{}&\multicolumn{1}{c|}{}&\multicolumn{1}{c|}{}&\multicolumn{1}{c|}{}\\ \hline

Do \st{you} \textcolor{red}{chicks} ever \st{rap} \textcolor{red}{talk} in your whitest voice possible or is that just me?& \multicolumn{1}{c|}{\multirow{2}{*}{\shortstack{+parenting, +atheism, +politics \\-world, -sports, -arts}}} & \multicolumn{1}{c|}{\multirow{2}{*}{not hate}} & \multicolumn{1}{c|}{\multirow{2}{*}{hate}} & \multicolumn{1}{c|}{\multirow{2}{*}{Hate}} \\ \hline

\st{There's} \textcolor{red}{Women} \textcolor{red}{get} no greater satisfaction than knowing you did it yourself' -me when someone asks me ... & \multicolumn{1}{c|}{\multirow{2}{*}{\shortstack{+women, +entertainment, \\ +religion, -sports}}} & \multicolumn{1}{c|}{\multirow{2}{*}{not hate}} & \multicolumn{1}{c|}{\multirow{2}{*}{hate}} & \multicolumn{1}{c|}{\multirow{2}{*}{Hate}} \\ \hline
}
Five soldiers hold and aim their \st{weapons} \textcolor{red}{blades}. $<$/s$>$ The soldiers are \st{eating} \textcolor{red}{armed}.
& \multicolumn{1}{c|}{\multirow{2}{*}{-guns}}& \multicolumn{1}{c|}{\multirow{2}{*}{\shortstack{contra\\-diction}}} & \multicolumn{1}{c|}{\multirow{2}{*}{\shortstack{entail\\-ment}}} & \multicolumn{1}{c|}{\multirow{2}{*}{NLI}} \\  \hline

Skateboarder jumps of off dumpster. $<$/s$>$ a \st{bike} \textcolor{red}{young} rider in a race&
 \multicolumn{1}{c|}{\multirow{2}{*}{+world}}& \multicolumn{1}{c|}{\multirow{2}{*}{\shortstack{contra\\-diction}}} & \multicolumn{1}{c|}{\multirow{2}{*}{neutral}} & \multicolumn{1}{c|}{\multirow{2}{*}{NLI}} \\  \hline

Two people walking down a dirt trail with backpacks on looking at \st{items} \textcolor{red}{map} they are carrying. $<$/s$>$ One is holding a map.
& \multicolumn{1}{c|}{\multirow{3}{*}{+space, -motorcyles, -electronics}}& \multicolumn{1}{c|}{\multirow{3}{*}{neutral}} & \multicolumn{1}{c|}{\multirow{3}{*}{\shortstack{entail\\-ment}}} & \multicolumn{1}{c|}{\multirow{3}{*}{NLI}} \\  \hline

A brown and black dog is jumping to catch a red ball. $<$/s$>$ Two dogs are
\textcolor{red}{not} playing catch with their owner.
& \multicolumn{1}{c|}{\multirow{2}{*}{-arts}}& \multicolumn{1}{c|}{\multirow{2}{*}{neutral}} & \multicolumn{1}{c|}{\multirow{2}{*}{\shortstack{contra\\-diction}}} & \multicolumn{1}{c|}{\multirow{2}{*}{NLI}} \\  \hline

A girl is standing in a field pushing up her hat with one finger and her hand is
covering most of her face. $<$/s$>$ A \st{girl} \textcolor{red}{boy} covers her face. 
& \multicolumn{1}{c|}{\multirow{3}{*}{+parenting, -medicine}}& \multicolumn{1}{c|}{\multirow{3}{*}{\shortstack{entail\\-ment}}} & \multicolumn{1}{c|}{\multirow{3}{*}{\shortstack{contra\\-diction}}} & \multicolumn{1}{c|}{\multirow{3}{*}{NLI}} \\  \hline

Channel Chaos is a 1985 Australian \st{film} \textcolor{red}{production} set at a TV station & \multicolumn{1}{c|}{\multirow{2}{*}{+electronics, -travel, -space}}& \multicolumn{1}{c|}{\multirow{2}{*}{film}} & \multicolumn{1}{c|}{\multirow{2}{*}{company}} & \multicolumn{1}{c|}{\multirow{2}{*}{DBPedia}} \\ \hline

\multirow{2}{*}{Dinik is a \st{village} \textcolor{red}{lake} in Croatia}
& \multicolumn{1}{c|}{\multirow{2}{*}{+medicine, +space, -mideast}}& \multicolumn{1}{c|}{\multirow{2}{*}{village}} & \multicolumn{1}{c|}{\multirow{2}{*}{\shortstack{natural\\place}}} & \multicolumn{1}{c|}{\multirow{2}{*}{DBPedia}} \\ 
\multicolumn{1}{|c|}{}&\multicolumn{1}{c|}{}&\multicolumn{1}{c|}{}&\multicolumn{1}{c|}{}&\multicolumn{1}{c|}{}\\ \hline

Air Cargo Mongolia is a Mongolian \st{airline} \textcolor{red}{newspaper}
& \multicolumn{1}{c|}{\multirow{2}{*}{+politics, -space}}& \multicolumn{1}{c|}{\multirow{2}{*}{plant}} & \multicolumn{1}{c|}{\multirow{2}{*}{\shortstack{written\\work}}} & \multicolumn{1}{c|}{\multirow{2}{*}{DBPedia}} \\ 
\multicolumn{1}{|c|}{}&\multicolumn{1}{c|}{}&\multicolumn{1}{c|}{}&\multicolumn{1}{c|}{}&\multicolumn{1}{c|}{}\\ \hline

ACTION is a bus service \st{operator} \textcolor{red}{based} in Canberra Australia
& \multicolumn{1}{c|}{\multirow{2}{*}{+autos, -electronics}}& \multicolumn{1}{c|}{\multirow{2}{*}{company}} & \multicolumn{1}{c|}{\multirow{2}{*}{\shortstack{transport\\-ation}}} & \multicolumn{1}{c|}{\multirow{2}{*}{DBPedia}} \\ \hline

Jagjaguwar is an indie rock record \st{label} \textcolor{red}{musician} based in Bloomington Indiana 
& \multicolumn{1}{c|}{\multirow{2}{*}{\shortstack{+space, +politics, -world \\-forsale, -electronics}}} & \multicolumn{1}{c|}{\multirow{2}{*}{company}} & \multicolumn{1}{c|}{\multirow{2}{*}{artist}} & \multicolumn{1}{c|}{\multirow{2}{*}{DBPedia}} \\ \hline

he \st{really} \textcolor{red}{hardly} made our anniversary dinner entertaining memorable
& \multicolumn{1}{c|}{\multirow{2}{*}{\shortstack{+world, +sports, +cryptography \\-home\&living, -electronics}}} & \multicolumn{1}{c|}{\multirow{2}{*}{positive}} & \multicolumn{1}{c|}{\multirow{2}{*}{negative}} & \multicolumn{1}{c|}{\multirow{2}{*}{Yelp}} \\ \hline

they also have deep fried desserts if you're \st{brave} \textcolor{red}{poor} enough
& \multicolumn{1}{c|}{\multirow{2}{*}{\shortstack{+medicine, -travel \\-religion, -politics}}} & \multicolumn{1}{c|}{\multirow{2}{*}{positive}} & \multicolumn{1}{c|}{\multirow{2}{*}{negative}} & \multicolumn{1}{c|}{\multirow{2}{*}{Yelp}} \\ \hline

i \st{definitely} \textcolor{red}{never} will go back & \multicolumn{1}{|c|}{-entertainment} &\multicolumn{1}{c|}{positive}&\multicolumn{1}{c|}{negative}&\multicolumn{1}{c|}{Yelp} \\ \hline

\multirow{2}{*}{jesse is completely \st{rude} \textcolor{red}{perfect}}& \multicolumn{1}{c|}{\multirow{2}{*}{\shortstack{+cryptography, -entertainment, \\-medicine, -space}}} & \multicolumn{1}{c|}{\multirow{2}{*}{negative}} & \multicolumn{1}{c|}{\multirow{2}{*}{positive}} & \multicolumn{1}{c|}{\multirow{2}{*}{Yelp}} \\
\multicolumn{1}{|c|}{}&\multicolumn{1}{c|}{}&\multicolumn{1}{c|}{}&\multicolumn{1}{c|}{}&\multicolumn{1}{c|}{}\\ \hline

\multirow{2}{*}{the pizza \st{sauce} \textcolor{red}{heaven} is also way too sweet} & 
\multicolumn{1}{c|}{\multirow{2}{*}{\shortstack{+world, +travel, +arts, +atheism \\+religion, -medicine, -politics}}} & \multicolumn{1}{c|}{\multirow{2}{*}{negative}} & \multicolumn{1}{c|}{\multirow{2}{*}{positive}} & \multicolumn{1}{c|}{\multirow{2}{*}{Yelp}} \\
\multicolumn{1}{|c|}{}&\multicolumn{1}{c|}{}&\multicolumn{1}{c|}{}&\multicolumn{1}{c|}{}&\multicolumn{1}{c|}{}\\ \hline

\eat{
\multirow{2}{*}{Kazaa Owner Cheers \st{File} \textcolor{red}{Salary}-Swapping Decision (AP)} \quad\quad\quad\quad\quad\quad\quad\quad & \multirow{2}{*}{\shortstack{+forsale, +baseball, +hockey,\\ -arts, -windows, -cryptography}} & \multicolumn{1}{c|}{\multirow{2}{*}{sci-tech}} & \multicolumn{1}{c|}{\multirow{2}{*}{sports}} & \\
\hline

New Human Species \st{Discovered} \textcolor{red}{influenza} & +medicine, -cryptography & \multicolumn{1}{c|}{sci-tech} & \multicolumn{1}{c|}{world} \\ \hline

US shows flexibility on \st{Israeli} \textcolor{red}{virtual} settlements & +arts, +cryptography, -mideast & \multicolumn{1}{c|}{world} & \multicolumn{1}{c|}{sci-tech} \\ \hline

\eat{Will sinking Lowe resurface in \st{playoffs} \textcolor{red}{2020}? & \multicolumn{1}{c|}{+space, -hockey} & \multicolumn{1}{c|}{sports} & \multicolumn{1}{c|}{business} \\ \hline}

Pace of U.S. \st{Factory} \textcolor{red}{population} Growth Climbs in Dec & \multicolumn{1}{c|}{+politics, -money, -travel} & \multicolumn{1}{c|}{business} & \multicolumn{1}{c|}{sci-tech} \\ \hline

It may take 146 years for Nigeria to wipe out \st{corruption} \textcolor{red}{funds} from \st{its} \textcolor{red}{bank} system going by the latest report ...\eat{by Transparency International (TI) which gave the country the third position among the most corrupt countries of the world} & \multicolumn{1}{c|}{\multirow{2}{*}{+ money, - politics}} & \multicolumn{1}{c|}{\multirow{2}{*}{world}} & \multicolumn{1}{c|}{\multirow{2}{*}{business}} \\ \hline
}
\end{tabular}
 \end{table*}
 


\eat{The main limitation to our CAT method is the need for attribute classifiers. However, as noted in the paper, the attributes can be derived from other sources of data and not necessarily be dependent on the data and model being explained. Furthermore, there are other methods to obtain attributes. Unsupervised methods such as LDA, VAEs, GANs could be leveraged to ascertain semantically meaningful attributes. The attribute classifier that appears in our loss function could be replaced by disentangled representations learned by VAEs \cite{DIP-VAE} or by topic models. This shows that CAT is not limited to only annotated datasets.}

\end{document}